\documentclass[11pt,a4paper]{article}
\usepackage[hyperref]{eacl2021}
\usepackage{times}
\usepackage{latexsym}

\usepackage{microtype}

\aclfinalcopy 
\usepackage{multirow}
\usepackage{multicol}

\usepackage{graphicx}  
\usepackage{booktabs}  

\usepackage{adjustbox}  

\usepackage{float}
\restylefloat{table}

\definecolor{msgrblue}{HTML}{4889f4}
\definecolor{msgrpaleblue}{HTML}{a9c6f5}
\definecolor{msgrgray}{HTML}{e1e1e7}
\newcommand{\lose}[1]{{\colorbox{msgrgray}{#1}}}
\newcommand{\tie}[1]{{\colorbox{msgrpaleblue}{#1}}}
\newcommand{\win}[1]{{\colorbox{msgrblue}{\color{white}{\textbf{#1}}}}}

\title{
Multi-Modal Open-Domain Dialogue
}

\author{Kurt Shuster\thanks{\,\,Joint First Authors.}, Eric Michael Smith$^*$, Da Ju, Jason Weston \\
\texttt{\{kshuster,ems,daju,jase\}@fb.com}
 }

\date{}

\begin{document}
\maketitle
\begin{abstract}
Recent work in open-domain conversational agents has demonstrated that significant improvements in model engagingness and humanness metrics can be achieved via massive scaling in both pre-training data and model size \cite{adiwardana2020meena,roller2020recipes}. However, if we want to build agents with human-like abilities, we must expand beyond handling just text. A particularly important topic is the ability to see images and  communicate about what is perceived. With the goal of engaging humans in multi-modal dialogue, we investigate combining components from state-of-the-art open-domain dialogue agents with those from state-of-the-art vision models. We study incorporating different image fusion schemes and domain-adaptive pre-training and fine-tuning strategies, and show that our best resulting model outperforms strong existing models in multi-modal dialogue while simultaneously performing as well as its predecessor (text-only) BlenderBot \cite{roller2020recipes} in text-based conversation.
We additionally investigate and incorporate safety components in our final model, and show that such efforts do not diminish model performance with respect to engagingness metrics.

\end{abstract}

\section{Introduction}

\begin{figure}[t!]
\centering
\includegraphics[width=6.5cm,height=6.5cm]{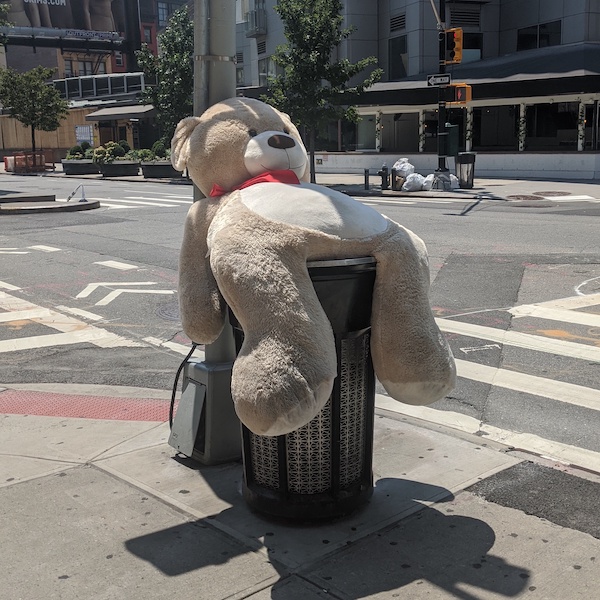}
\includegraphics[width=0.48\textwidth]{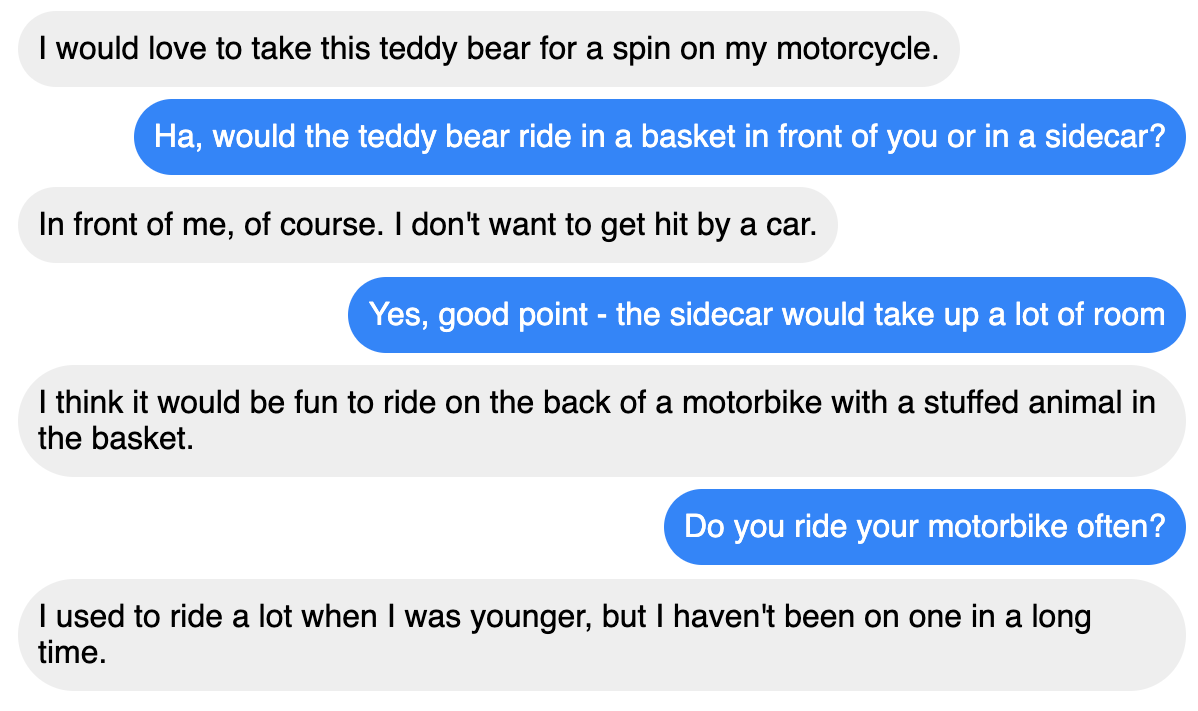}
\caption{
Paper author (right speaker) conversing with our MMB DegenPos model (left speaker). This example was cherry picked. We show more sample conversations in Section~\ref{sec:removing_style}.
 \label{fig:cherry_convos1}
}
\end{figure}

An important goal of artificial intelligence is the construction of open-domain conversational agents that can engage humans in discourse. Indeed, the future of human interaction with AI is predicated on models that can exhibit a number of different conversational skills over the course of rich dialogue. Much recent work has explored building and training dialogue agents that can blend such skills throughout natural conversation, with the ultimate goal of providing an engaging and interesting experience for humans \cite{smith2020bst, shuster2019dialogue}. Coupled with the advancement of large-scale model training schemes, such models are becoming increasingly engaging and human-like when compared to humans \cite{zhang2019dialogpt,adiwardana2020meena, roller2020recipes}.

In order to better approach human-like ability, however,
it is necessary that agents can converse with both textual and visual context, similarly to how humans interact in the real world; 
indeed, communication grounded in images is naturally engaging to humans \cite{hu2014we}. Recent efforts have gone beyond classical, fact-based tasks such as image captioning or visual question answering \cite{antol2015vqa,das2017visual} to produce models that can respond and communicate about images in the flow of natural conversation \cite{shuster2020image,shuster2019dialogue}.

In this work, we explore the extension of large-scale conversational agents to image-based dialogue. We combine representations from image-based models that have been trained on object detection tasks \cite{Lu_2020_CVPR, NIPS2019_8297} with representations from Transformers with billions of parameters pre-trained on massive (text-only) dialogue datasets, to produce 
responses conditioned on both visual and textual context. To ensure that our model retains the ability to engage in regular, text-based conversation, we include in our training procedure multi-tasking with datasets expressly designed to instill conversational skills in the model \cite{smith2020bst}. 

We find that our best resulting models are as proficient in text-only conversation as the current best reported dialogue models, with respect to both automated metrics measuring performance on the relevant datasets and human evaluations of engagingness.
Simultaneously, our model significantly outperforms recent strong multi-modal dialogue models when in an image-dialogue regime; we measure several metrics via pairwise human judgments using ACUTE-Eval \cite{li2019acute} to show that our model is not only more engaging but can also discuss and reference visual context throughout a conversation. 
See Figure~\ref{fig:cherry_convos1} for one sample cherry-picked conversation with our model, with random and lemon-picked conversations in Figures~\ref{fig:cherry_convos2} and \ref{fig:lemon_convos}.

One important avenue that we explore with our best models is safety - that is, ensuring that our models are not offensive 
to their conversational partners.
Dialogue safety is indeed a well-studied, but still unsolved, research area
\cite{dinan2019safety,liu2019does,dinan2019queens,blodgett-etal-2020-language,ch2018detecting,schafer-burtenshaw-2019-offence, zhang-etal-2018-conversations-awry}, yet we note that safety in the context of image-dialogue is relatively less explored. 
In this work we examine gender bias and toxicity of text generations in the context of various  styles from the Image-Chat dataset \cite{shuster2020image}. Notably, after tuning the model to reduce toxicity and gender bias, we find that model engagingness does not diminish. 

We make publicly available the training procedure, initial pre-trained model weights, and datasets in ParlAI \footnote{\url{https://github.com/facebookresearch/ParlAI/tree/master/projects/multimodal_blenderbot}} to allow for fully reproducible results. 
%

\section{Related Work}

\subsection{Multi-Modal Models and Tasks}

\paragraph{Rich Representations} 
Modeling multi-modal inputs, i.e. in the visual + textual context, is a well-researched area. Much of the existing literature explores similar architectures to our setup, i.e., using standard Transformer-based models to jointly encode text and images \cite{li2019visualbert, kiela2019supervised}. Others have explored modifications to the standard self-attention scheme in Transformers by incorporating additional co-attention \cite{NIPS2019_8297, tan-bansal-2019-lxmert} or cross-attention \cite{stefanini2020novel} layers. These models have primarily been used for generating rich joint representations of images and text for use in downstream tasks, and they primarily focus on the encoding aspect. 

\paragraph{Visual Dialogue/Caption Generation} Many tasks have been designed to measure the ability of a model to produce text in the context of images. Specifically, COCO Captions \cite{chen2015microsoft} and Flickr30k \cite{young2014image} require a model to produce a caption for a given image. A variety of sequence-to-sequence \cite{vinyals2015show,xu2015show,anderson2017bottom} and retrieval-based \cite{8578848,faghri2018vse++,Nam2016DualAN} models have been applied to these tasks, however they do not go beyond the one-turn text generation expected for captioning an image. Other recent architectures have explored text generation \cite{wang2020vdbert, park2020multiview} in the context of the Visual Dialog \cite{visdial} task; however, this task is primarily used to measure a model's ability to answer questions about an image in the flow of a natural conversation, which differs somewhat from the open-domain dialogue task. Further still, there have been recent forays into open-domain natural dialogue in the context of images, e.g. in the Image-Chat \cite{shuster2020image} and Image-grounded Conversations \cite{mostafazadeh-etal-2017-image} tasks. Again, retrieval-based \cite{shuster2020image,ju2019allinone} and sequence-to-sequence \cite{shuster2019dialogue, shuster2020image} models have been used to conduct dialogue in this regime. 





\subsection{Multi-Task Training / Using Pre-Trained Representations}
Our multi-modal model is constructed from models pre-trained in other, related domains; specifically, we seek to fuse the resulting weights of large-scale, uni-modal pre-training to achieve good performance on downstream, multi-modal tasks. Adapting pre-trained representations to later downstream tasks has been shown to be successful in NLP \cite{Peters_2019, devlin2019bert} and dialogue in particular \cite{roller2020recipes,mazare2018trainingmillions}. Additionally, large-scale multi-modal pre-training has been shown to be effective in other downstream multi-modal tasks \cite{unicodervl, chen2020uniter,singh2020pretraining}. Our work does not contain multi-modal pre-training in itself, but rather we explore what some have deemed ``domain-adaptive pre-training'' \cite{Gururangan_2020} or ``intermediate task transfer" \cite{intermediatetasktransfer}, in which pre-trained representations are ``adapted" to a certain domain via an intermediate training step, before training/evaluating on the requisite downstream tasks. It is also worth noting that we employ multi-task training, to both help generalize the applicability of the model and improve its performance on downstream tasks/evaluations; this has been shown recently to help in both image-based \cite{singh2020pretraining,ju2019allinone,Lu_2020_CVPR} and text-based \cite{shuster2019dialogue, roller2020recipes} tasks. 

\subsection{Comparison to Existing Models}
\label{sec:existing_models_related_work}

In this work, we compare our best resulting model to the following existing models in the literature:

\begin{itemize}
  \item The 2.7-billion-parameter Transformer sequence-to-sequence model from \citet{roller2020recipes}, known as ``BST Generative 2.7B model'' in that work, pre-trained on 1.5B comments from a third-party Reddit dump hosted by pushshift.io \cite{baumgartner2020pushshift}. We refer to this model as ``BlenderBot''.
  \item DialoGPT, a GPT-2-based model trained on 147M exchanges from public-domain social-media conversations \citep{zhang2019dialogpt}
  \item Meena, a 2.6B-parameter Transformer sequence-to-sequence model trained on 341GB of conversations \citep{adiwardana2020meena}
  \item The Image+Seq2Seq model from dodecaDialogue \citep{shuster2019dialogue}, a Transformer sequence-to-sequence model in which the encoder is passed pre-trained image features from the ResNeXt-IG-3.5B model \citep{uru}. We use the dodecaDialogue model fine-tuned on Image-Chat (and we refer to this model as ``Dodeca'').
  \item 2AMMC \citep{ju2019allinone}, in which multiple Transformers are attended over in order to make use of a combination of ResNeXt-IG-3.5B and Faster R-CNN image features \citep{Detectron2018}. We specifically use the 2AMMC model from \citet{ju2019allinone} because that model has the best test-set performance on Image-Chat in that work.
\end{itemize}

\section{Model Architectures}
The inputs to our models are visual and/or textual context, where applicable. We explore different ways to encode images from their pixels to vector representations, and we additionally compare ways of combining (fusing) the image and text representations before outputting a response.
\subsection{Image Encoders}
Converting an image from pixels to a vector representation is a well-researched problem, and thus we explore using two different image encoders to determine the best fit for our tasks.
\begin{itemize}
    \item \textbf{ResNeXt WSL} We first experiment with image representations obtained from pre-training a ResNeXt 32x48d model on nearly 1 billion public images \cite{uru}, with subsequent fine-tuning on the ImageNet1K dataset \cite{imagenet} \footnote{\url{https://pytorch.org/hub/facebookresearch_WSL-Images_resnext/}}. The output of this model is a 2048-dimensional vector, and we refer to these representations as ``ResNeXt WSL" features. 
    \item \textbf{ResNeXt WSL Spatial} One can also take the output of the image encoder prior to its final fully-connected layer to obtain ``spatial" image features, resulting in a 2048$\times$7$\times$7-dimensional vector. We explore results with these features as well, and refer to them as ``ResNeXt WSL Spatial".
    \item \textbf{Faster R-CNN} Finally, we consider Faster R-CNN features \cite{Ren_2017}, using models trained in the Detectron framework \cite{Detectron2018}; specifically, we use a ResNeXt-152 backbone trained on the Visual Genome dataset \cite{krishnavisualgenome} with the attribute head \cite{singh2020mmf} \footnote{\url{https://github.com/facebookresearch/vilbert-multi-task}}. The Faster R-CNN features are 2048$\times$100-dimensional representations, and we refer to these features as ``Faster R-CNN". 
\end{itemize}

\subsection{Multi-Modal Architecture}
To jointly encode visual and textual context, we use a modification of a standard Transformer sequence-to-sequence architecture \cite{vaswani2017attention}, whereby we experiment with different ways of combining (fusing) the image and text representations to generate an output sequence. Our Transformer model has 2 encoder layers, 24 decoder layers, 2560-dimensional embeddings, and 32 attention heads, and the weights are initialized from a 2.7-billion parameter model pre-trained on 1.5B comments from a third-party Reddit dump hosted by pushshift.io \cite{baumgartner2020pushshift} to generate a comment conditioned on the full thread leading up to the comment \cite{roller2020recipes}. From this base model, we explore two possible fusion schemes. 
\paragraph{Late Fusion} The late fusion method is the same as used in \citet{shuster2019dialogue}, whereby the encoded image is projected to the same dimension as the text encoding of the Transformer encoder, concatenated with this output as an extra ``token" output, and finally fed together as input to the decoder.
\paragraph{Early Fusion}
We additionally experiment with an earlier fusion scheme to allow greater interaction between the image and text in the sequence-to-sequence architecture. In a similar fashion to VisualBERT \cite{li2019visualbert} and multi-modal Bitransformers \cite{kiela2019supervised}, we concatenate the projected image encoding from the visual input with the token embeddings from the textual input, assign each a different segment embedding, and jointly encode the text and image in the encoder. The encoder thus performs full self-attention across the textual and visual context, with the entire output used as normal in the sequence-to-sequence architecture.

As our resulting model can be seen as a multi-modal extension to the BlenderBot model \cite{roller2020recipes}, we refer to it as ``Multi-Modal BlenderBot" (MMB).

\section{Training Details}
When training the model, we fix the weights of the pre-trained image encoders, except the linear projection to the Transformer output dimension, and fine-tune all of the weights of the Transformer encoder/decoder.
\subsection{Domain-Adaptive Pre-Training}
During training, the vast majority of trainable model weights are initialized from a large, 2.7B parameter Transformer pre-trained solely on textual input. As our end goal is to achieve improved performance on multi-modal tasks, we found that training first on domain-specific/related data was helpful in order to adapt the Transformer model to an image setting. Following \cite{singh2020pretraining}, we experimented with pre-training on COCO Captions \cite{chen2015microsoft} - a dataset of over 120k images with 5 captions each, resulting in over 600k utterances - in which the model is trained to generate a caption solely from image input. We additionally explored multi-tasked training with COCO Captions and on the same third-party Reddit dump hosted by pushshift.io \cite{baumgartner2020pushshift} as the one used in pre-training the Transformer model, to see whether it was necessary to ensure the model did not stray too far from its ability to handle pure textual input.

During domain-adaptive pre-training, we trained the model on 8 
GPUs for 10k-30k SGD updates, using early-stopping on the validation set. The models were optimized using Adam \cite{kingma2014adam}, with sweeps over a learning rate between 5e-6 and 3e-5, using 100 warmup steps.

\subsection{Fine-tuning Datasets}
\label{sec:fine-tune-datasets}
The goal of our resulting model is to perform well in a multi-modal dialogue setting; thus, we fine-tune the model on both dialogue and image-dialogue datasets. For dialogue-based datasets, we consider the same four as in \citet{roller2020recipes}: ConvAI2 \cite{dinan2019second}, EmpatheticDialogues \cite{rashkin2019empathetic}, Wizard of Wikipedia \cite{dinan2018wizard}, and BlendedSkillTalk  \cite{smith2020bst}. To model image-dialogue, we consider the Image-Chat dataset \cite{shuster2020image}. We give a brief description below of the five datasets; more information can be found in \citet{roller2020recipes} and \citet{shuster2020image}.

\paragraph{ConvAI2}
The ConvAI2 dataset \cite{dinan2019second} is based on the Persona-Chat \cite{zhang2018personalizing} dataset, and contains 140k training utterances in which crowdworkers were given prepared ``persona" lines, e.g. ``I like dogs" or ``I play basketball", and then paired up and asked to get to know each other through conversation.
\paragraph{EmpatheticDialogues (ED)} The EmpatheticDialogues dataset \cite{rashkin2019empathetic} was created via crowdworkers as well, and involves two speakers playing different roles in a conversation. One is a ``listener", who displays empathy in a conversation while conversing with someone who is describing a personal situation. The model is trained to act like the ``listener". The resulting dataset contains 50k utterances.
\paragraph{Wizard of Wikipedia (WoW)} The Wizard of Wikipedia dataset \cite{dinan2018wizard} involves two speakers discussing a given topic in depth, comprising 194k utterances. One speaker (the ``apprentice'') attempts to dive deep on and learn about a chosen topic; the other (the ``wizard'') has access to a retrieval system over Wikipedia, and is tasked with teaching their conversational partner about a topic via grounding their responses in a knowledge source. 
\paragraph{BlendedSkillTalk (BST)} BlendedSkillTalk \cite{smith2020bst} is a dataset that essentially combines the three above. That is, crowdworkers are paired up similarly to the three previous datasets, but now all three ``skills" (personalization, empathy, and knowledge) are at play throughout the dialogue: the speakers are tasked with blending the skills while engaging their partners in conversation. The resulting dataset contains 74k utterances.
\paragraph{Image-Chat (IC)} The Image-Chat dataset \cite{shuster2020image} contains 200k dialogues over 200k images: crowdworkers were tasked with discussing an image in the context of a given style, e.g. ``Happy'', ``Cheerful'', or ``Sad'', in order to hold an engaging conversation. The resulting dataset contains over 400k utterances. For each conversation in the dataset, the two speakers are each assigned a style in which that speaker responds, and these styles are optionally fed into models as part of the input, alongside the dialogue context. There are 215 styles in total, and styles are divided into 3 categories, ``positive'', ``neutral'', and ``negative''.\footnote{Lists of positive, neutral, and negative styles are from \url{http://ideonomy.mit.edu/essays/traits.html}, following \citet{shuster2018imagecaption}.}

\paragraph{} In the fine-tuning stage, we consider two different regimes: one in which we multi-task train on the five datasets together, and one in which we train on Image-Chat alone. While the latter regime is useful in exploring upper bounds of model performance, our main goal is to build a model that can display the requisite skills of an engaging conversationalist (empathy, personalization, knowledge) while \textit{also} having the ability to respond to and converse about images; thus, we are more interested in the former training setup. In this stage, we train the models on 8 GPUs for around 10k train updates using a similar optimization setup as in the domain-adaptive pre-training stage. 

\section{Experiments}

\subsection{Automatic Evaluations}

\begin{table*}[t!]
\begin{center}
\small
\begin{tabular}{|l|l|r|r|r|}
 \hline
Image Features & Image Fusion & COCO (ppl) & pushshift.io Reddit (ppl) & Average\\
\hline
\multicolumn{5}{c}{COCO \& pushshift.io Reddit training data} \\
\hline
ResNeXt WSL & Late & 11.11 & 13.80 & 12.45\\
 & Early & 6.69 & 13.50 & 10.10\\
\hline
ResNeXt WSL Spatial & Late & 7.43 & 13.00 & 10.22 \\
 & Early & 6.53 & 13.46 & 10.00\\
\hline
Faster R-CNN & Late & 5.26 & 13.17 & 9.21 \\
 & Early & 5.23 & 13.15 & \textbf{9.13}\\
\hline
\multicolumn{5}{c}{COCO training data only} \\
\hline
ResNeXt WSL & Late & 5.82 & 19.52 & 12.67 \\
 & Early & 6.21 & 21.30 & 13.76\\
\hline
ResNeXt WSL Spatial & Late & 6.51 & 16.50 & 11.51 \\
 & Early & 6.19 & 18.77 & 12.48 \\
\hline
Faster R-CNN & Late & 5.21 & 17.88 & 11.55\\
 & Early & \textbf{4.83} & 18.81 & 11.82 \\
\hline
\end{tabular}
\caption{Model performance, measured via perplexity on validation data, on domain-adaptive pre-training datasets, comparing various image features and image fusion techniques. The top three rows involve multi-task training on COCO Captions and pushshift.io Reddit, while the bottom three rows involve single task training on COCO Captions only. We note that early fusion with Faster R-CNN features yields the best performance on COCO Captions.
\label{table:pre_training_results}
}
\end{center}
\end{table*}

\begin{table*}[t!]
\begin{center}
\small
\scriptsize
\begin{tabular}{|l|l|l|r|r|r|r|r|r||r|r| }
 \hline
Image  & Training & Image & ConvAI2 & ED & WoW & BST & IC 1st & IC & Text & All \\
Features & Data & Fusion & & &  & & Turn  &  & Avg. & Avg. \\
\hline
  & None & & 12.31 & 10.21 & 13.00 & 12.41 & 32.36 & 21.48 & 11.98 & 13.88 \\
None & BST$^+$ & None & 8.74 & 8.32 & 8.78 & 10.08 & 38.94 & 23.13 & 8.98 & 14.76 \\ 
 & BST$^+$ + IC & & 8.72 & 8.24 & 8.81 & 10.03 & 16.03 & 13.21 & \textbf{8.95} & 9.83 \\ 
 \hline
 \hline
 & BST$^+$ + IC & Late & 8.71 & 8.25 & 8.87 & 10.09 & 16.20 & 13.27 & 8.98 & 9.84 \\ 
 & BST$^+$ + IC & Early  & 8.80 & 8.32 & 8.79 & 10.17 & 15.16 & 12.99 & 9.02 & 9.81 \\ 
ResNeXt WSL & BST$^+$ + IC + COCO + Reddit & Late & 9.27 & 8.87 & 9.45 & 10.74 & 17.56 & 14.44 & 9.58 & 10.56 \\ 
 & BST$^+$ + IC + COCO + Reddit & Early & 9.34 & 8.90 & 9.48 & 10.78 & 15.87 & 13.88 & 9.62 & 10.48 \\ 
 & BST$^+$ + IC + COCO & Late & 8.79 & 8.36 & 9.00 & 10.21 & 16.00 & 13.31 & 9.09 & 9.93 \\ 
 & BST$^+$ + IC + COCO & Early & 8.91 & 8.38 & 8.99 & 10.29 & 14.64 & 12.85 & 9.14 & 9.88 \\ 
 
 \hline
 \hline
 & BST$^+$ + IC& Late & 8.71 & 8.24 & 8.88 & 10.10 & 15.39 & 13.02 & 8.98 & 9.78 \\ 
 & BST$^+$ + IC& Early& 8.79 & 8.29 & 8.92 & 10.15 & 15.34 & 13.02 & 9.04 & 9.83 \\ 
ResNeXt WSL & BST$^+$ + IC + COCO + Reddit& Late & 8.76 & 8.31 & 8.88 & 10.14 & 15.20 & 13.04 & 9.02 & 9.83 \\  
Spatial & BST$^+$ + IC + COCO + Reddit& Early& 9.30 & 8.82 & 9.46 & 10.76 & 15.67 & 13.79 & 9.56 & 10.43 \\ 
 & BST$^+$ + IC + COCO& Late & 8.73 & 8.31 & 8.87 & 10.13 & 15.04 & 12.98 & 9.01 & 9.84 \\ 
 & BST$^+$ + IC + COCO& Early& 8.81 & 8.34 & 8.99 & 10.22 & 14.76 & 12.87 & 9.09 & 9.80 \\ 

 \hline
 \hline
 & BST$^+$ + IC & Late & 8.70 & 8,24 & 8.92 & 10.07 & 13.97 & 12.48 & 8.98 & \textbf{9.68} \\ 
 & BST$^+$ + IC & Early& 8.81 & 8.33 & 8.81 & 10.15 & 13.66 & 12.43 & 9.03 & 9.71 \\ 
 & BST$^+$ + IC + COCO + Reddit& Late & 8.75 & 8.31 & 8.93 & 10.14 & 13.83 & 12.49 & 9.03 & 9.73 \\  
 Faster R-CNN & BST$^+$ + IC + COCO + Reddit& Early& 8.78 & 8.31 & 8.85 & 10.15 & \textbf{13.51} & \textbf{12.36} & 9.02 & 9.69 \\ 
& BST$^+$ + IC + COCO& Late & 8.74 & 8.33 & 8.87 & 10.13 & 13.85 & 12.51 & 9.02 & 9.72 \\
 & BST$^+$ + IC + COCO& Early& 8.81 & 8.34 & 8.93 & 10.19 & 13.57 & 12.39 & 9.07 & 9.73 \\ 
 
 \hline
\end{tabular}
\caption{Ablation analysis of the impact of various image features, training data (including domain-adaptive pre-training), and image fusion techniques on the datasets described in Section \ref{sec:fine-tune-datasets}, where BST$^+$ refers to the four text-only dialogue datasets (ConvAI2, ED, WoW, and BST). The numbers shown are model perplexities measured on each of the datasets' validation data. Performance on the first turn of Image-Chat is also measured to highlight model performance when only given visual context. We note that using Faster R-CNN image features results in the best average performance, as well as the best performance on Image-Chat. 
\label{table:ablation_results}
}
\end{center}
\end{table*}

\begin{table*}[t!]
\begin{center}
\small
\begin{tabular}{|l|l|l|r|r| }
 \hline
Image  & Training & Image & IC First Turn & IC \\
Features & Data & Fusion & & \\
\hline
None & None & None & 32.36 & 21.48 \\ 
& Image Chat & & 28.71 & 13.17 \\ 
\hline
\hline
& IC & Late  &14.80 & 12.83 \\ 
& IC & Early  &16.00 & 13.21 \\ 
& IC + COCO + Reddit & Late & 16.73 & 13.92 \\ 
ResNeXt WSL & IC + COCO + Reddit & Early & 15.71 & 13.53 \\ 
& IC + COCO & Late & 14.70 & 12.95 \\ 
& IC + COCO & Early & 14.62 & 12.92 \\ 

\hline
\hline
& IC & Late  &15.34 & 13.01 \\ 
& IC & Early  &15.27 & 13.00 \\ 
ResNeXt WSL  & IC + COCO + Reddit & Late & 15.09 & 12.95 \\ 
 Spatial & IC + COCO + Reddit & Early & 15.55 & 13.50 \\ 
& IC + COCO & Late & 15.02 & 12.95 \\ 
& IC + COCO & Early & 14.62 & 12.87 \\ 

\hline
\hline
& IC & Late  &13.99 & 12.51 \\ 
& IC & Early  &13.76 & 12.42 \\ 
& IC + COCO + Reddit & Late & 13.75 & 12.43 \\ 
Faster R-CNN & IC + COCO + Reddit & Early & \textbf{13.44} & \textbf{12.29} \\ 
& IC + COCO & Late & 13.82 & 12.48 \\
& IC + COCO & Early & 13.56 & 12.37 \\ 

 \hline
\end{tabular}
\caption{Ablation analysis of the impacts of various image features, training data (including domain-adaptive pre-training), and image fusion techniques when training on the Image-Chat dataset alone (i.e., ignoring the text-only dialogue datasets). As in Table \ref{table:ablation_results}, we note that Faster R-CNN features yield the best results on Image-Chat.
\label{table:ablation_results_ic}
}
\end{center}
\end{table*}

\begin{table*}[t!]
\begin{center}
\small
\begin{tabular}{|l|r|r|r|r|}
 \hline
Dataset & \multicolumn{1}{c}{PPL} & \multicolumn{1}{|c|}{F1} & \multicolumn{1}{|c|}{BLEU-4} & \multicolumn{1}{|c|}{ROUGE-L} \\
\hline
Image Chat (First Round) & 13.56 & 11.96 & 0.411 & 16.72 \\
Image Chat & 12.64 & 13.14 & 0.418 & 18.00 \\
BlendedSkillTalk & 9.98 & 17.84 & 0.980 & 19.25 \\
Wizard of Wikipedia (Seen) & 8.82 & 18.63 & 2.224 & 17.39 \\
ConvAI2 & 8.78 & 18.41 & 1.080 & 22.64 \\
EmpatheticDialogues & 8.46 & 19.23 & 1.448 & 24.46 \\
\hline
\end{tabular}
\caption{Test results of best multi-task model on BST$^+$ and Image Chat datasets, measured via perplexity (ppl), F1, BLEU-4, and ROUGE-L scores. ConvAI2 results are reported on the validation set, as the test set is hidden.
\label{table:final_auto_results}
}
\end{center}
\end{table*}

\begin{table*}[t!]
\begin{center}
\small
\scriptsize
\begin{tabular}{|l|rrr|rrr|rrr|rrr|rrr|}
 \hline
Model & \multicolumn{3}{|c|}{ConvAI2} & \multicolumn{3}{|c|}{ED} & \multicolumn{3}{|c|}{WoW Seen} & \multicolumn{3}{|c|}{BST} & \multicolumn{3}{|c|}{IC} \\
& F1 & B & R & F1 & B & R & F1 & B & R & F1 & B & R & F1 & B & R \\
\hline
\hline
DialoGPT & 11.4 & 0.1 & 8.5 & 10.8 & 0.3 & 8.2 & 8.6 & 0.1 & 5.9 & 10.5 & 0.1 & 7.6 & 6.2 & 0.1 & 5.2 \\
\cite{zhang2019dialogpt} & & & & & & & & & & & & & & & \\
\hline
Dodeca & 21.7 & 5.5 & 33.7 & 19.3 & 3.7 & 31.4 & 38.4* & 21.0* & 45.4* & - & - & - & 12.9 & 2.1 & 24.6 \\
\cite{shuster2019dialogue}& & & & & & & & & & & & & & &\\
\hline
2AMMC & - & - & - & - & - & - & - & - & - & - & - & - & 9.3 & 0.1 & 11.0 \\
\cite{ju2019allinone} & & & & & & & & & & & & & & & \\
\hline
BlenderBot & 18.4 & 1.1 & 22.7 & 19.1 & 1.4 & 24.2 & 18.8 & 2.3 & 17.5 & 17.8 & 1.0 & 19.2 & 9.2 & 0.1 & 12.3 \\
\cite{roller2020recipes} & & & & & & & & & & & & & & & \\
\hline
\hline
Multi-Modal BlenderBot & 18.4 & 1.1 & 22.6 & 19.2 & 1.5 & 24.5 & 18.6 & 2.2 & 17.4 & 17.8 & 1.0 & 19.3 & 13.1 & 0.4 & 18.0 \\
(ours) & & & & & & & & & & & & & & & \\
\hline
\end{tabular}
\caption{Test performance of existing models on the datasets considered, compared to MMB (specifically, the ``MMB Style'' model discussed in Section~\ref{sec:human_eval_chat_no_image}), in terms of F1, BLEU-4 (B), and ROUGE-L (R) scores. * indicates that gold knowledge was utilized in the WoW task.
\label{table:final_auto_results_existing}
}
\end{center}
\end{table*}

\subsubsection{Results on Pre-Training Datasets}
To fully understand the effects of various training data and image features, as well as multi-modal fusion schemes, we measure model perplexity on the COCO and pushshift.io Reddit validation sets. We are primarily interested in performance on COCO Captions, as the model has already been extensively pre-trained on the pushshift.io Reddit data. The results are shown in Table \ref{table:pre_training_results}. 

\paragraph{Training Data} We first note that, regardless of image fusion and image feature choices, we see the best performance on COCO Captions by simply fine-tuning exclusively on that data. This is an expected result, though we do see that in nearly every scenario the decrease in perplexity is not large (e.g. 5.23 for Faster R-CNN early fusion multi-tasking, down to 4.83 with just COCO Captions).

\paragraph{Image Features} Across all training setups, we see that using spatially-based image features (ResNeXt WSL Spatial, Faster R-CNN) yields better performance than just a single vector image representation (ResNeXt WSL). This difference is particularly noticeable when training with COCO and pushshift.io Reddit, where with Faster R-CNN features the model obtains an average ppl of 9.13 over the two datasets, while with ResNeXt WSL features the model only obtains 10.1 ppl. We find that using Faster R-CNN features additionally outperforms using ResNeXt WSL Spatial features, where using the latter obtains an average of 10.0 ppl over the two datasets.

\paragraph{Image Fusion} Finally, holding all other variables constant, we find that using our early fusion scheme yields improvements over using a late fusion scheme. E.g., with Faster-R-CNN features in the COCO-only setup, we see a decrease in perplexity from 5.21 to 4.83; with ResNeXt WSL Spatial image features, we see perplexity differences ranging from 0.3 to 0.9 depending on the training data.

\subsubsection{Results on Fine-Tuned Datasets}
We conduct the same ablation setups for training on the dialogue and image-and-dialogue datasets as we did in the domain-adapative pre-training setup; the results for multi-tasking \textit{all} of the datasets are in Table \ref{table:ablation_results}, while results for fine-tuning on Image-Chat alone are in Table \ref{table:ablation_results_ic}.

From these extensive ablations, we note some interesting conclusions.

\paragraph{Text-Only Datasets}
First, we look at the performance of our models on the text-only datasets. The second-to-last column in Table \ref{table:ablation_results} shows the average perplexity across the text-only datasets. If we compare the model that performs best on Image-Chat across all sets of image features (Faster-R-CNN features with BST$^+$ + IC + COCO + Reddit training data with early fusion) to the model in row 2, which is trained both without images and without Image-Chat on the text-only datasets, we see that the perplexity differences are quite small: that is, including training on an image-dialogue dataset, and overloading the Transformer encoder/decoder to incorporate image features, \textit{does not} hinder dialogue performance.

\paragraph{Training Data}
Across all image-feature choices, we see that the choice of training data indeed makes a difference in performance on Image-Chat. Examining the early fusion model in Table \ref{table:ablation_results}, by including COCO Captions (and, in some cases, pushshift.io Reddit) in the training data we see drops in perplexity from 12.99 to 12.85, 13.02 to 12.87, and 12.43 to 12.36 with ResNeXt WSL, ResNeXt WSL Spatial, and Faster R-CNN features respectively. The decrease in perplexity indicates that domain-adaptive pre-training indeed improves performance on Image-Chat. This difference is highlighted even more when we measure performance on the first turn of Image-Chat, in which the model must generate a response given no textual context: 15.16 to 14.64, 15.34 to 14.76, and 13.66 to 13.51.
We note a similar trend in Table \ref{table:ablation_results_ic}.

\paragraph{Image Features}
Again, we see that using Faster R-CNN features leads to dramatic improvements compared to using the ResNeXt WSL features (spatial or otherwise), yielding 12.36 perplexity on Image-Chat compared to 12.85 and 12.87 perplexity with ResNeXt WSL (non-spatial and spatial respectively) during multi-tasking, and 12.29 perplexity on Image-Chat compared to 12.92 and 12.87 respectively for single-task training on Image-Chat (see Table \ref{table:ablation_results_ic}).

\paragraph{Image Fusion}
Finally, we note as before that using our early fusion technique improves performance on Image-Chat across all ablation regimes. While the average perplexity across the dialogue datasets is best when using late image fusion, we obtain the best image chat perplexity when performing early image fusion.

\paragraph{Final Test Results}
Following the ablation analyses, we decide to compare our best multi-tasked and single-tasked trained model (with respect to the fine-tuning datasets), where we use Faster R-CNN image features and an early fusion scheme, to existing models in the literature.  For this comparison, we consider additional metrics that can be computed on the actual model generations: F1, BLEU-4 and ROUGE-L. We generate model responses during inference with the same generation scheme as in \citet{roller2020recipes} - beam search with beam size of 10, minimum beam length of 20, and tri-gram blocking within the current generation and within the full textual context. The test performance of our best multitask model on the various datasets is shown in Table \ref{table:final_auto_results}, with comparisons to existing models from Section \ref{sec:existing_models_related_work} in Table \ref{table:final_auto_results_existing}; all evaluations are performed in ParlAI \footnote{\url{https://parl.ai}}.

We first note that the Dodeca model performs well across the board, and indeed has the highest ROUGE-L, BLEU-4, and F1 scores for the three text-only datasets. Higher BLEU-4 scores can be attributed to specifying a smaller minimum generation length, as forcing the BlenderBot models to generate no less than 20 tokens hurts precision when compared to reference labels - this was verified as we tried generating with a smaller minimum length (5 tokens) and saw a 20\% increase in BLEU-4 on Image-Chat for Multi-Modal BlenderBot. Higher ROUGE-L scores can additionally be attributed to specifying a \textit{larger} minimum generation length; this was also verified by generating with a higher minimum length (50 tokens) where we saw nearly a 40\% increase in ROUGE-L score.
Nevertheless, we do not report an exhaustive search over parameters here for our model, and instead compare it to BlenderBot with the same settings next.

When compared to its predecessor, text-only BlenderBot, MMB performs nearly the same on all four text-only datasets, indicating that MMB has not lost its proficiency in text-only dialogue. Additionally, when comparing performance on Image-Chat to models trained on multi-modal data, MMB outperforms Dodeca in terms of F1 score (13.1 vs. 12.9) and outperforms 2AMMC on all three metrics. For the 2AMMC model, these metrics are computed under the assumption that the model's chosen response (from a set of candidate responses collated from the Image-Chat training set) is the ``generated" response.

\subsection{Human Evaluations}

\subsubsection{Summary of Human Evaluations}

Since our model must demonstrate compelling performance both in general chit-chat dialogue and when responding to an image when conversing with humans, we present several types of human evaluations in this section. Section~\ref{sec:human_eval_chat_no_image} presents evaluations of chit-chat performance, Section~\ref{sec:human_eval_image_response} shows evaluations on how well the model responds to sample images, and Section~\ref{sec:human_eval_image_chat} combines both of these skills by demonstrating how well the model performs when talking to a human about an image.

\subsubsection{Human/Model Chats Without Images}
\label{sec:human_eval_chat_no_image}

\begin{table}[t!]
\begin{center}
\small
\centering
\begin{tabular}{r|rrr}
\hline
& MMB Style & BlenderBot \\
\hline
Contradiction & 2.15\% & 3.37\% \\
Improper English & 0.27\% & 0.26\% \\
Repetitive & 1.34\% & 1.55\% \\
Unrelated & 2.42\% & 2.33\% \\
Non-Sensical & 4.03\% & 2.07\% \\
None (All Good) & 91.13\% & 91.45\% \\
\hline
Mean engagingness & 4.70$\pm$0.60 & 4.70$\pm$0.60 \\
\hline
\end{tabular}
\caption{Per-turn annotations and mean engagingness ratings of human/model conversations with MMB Style and BlenderBot. Both models perform roughly equivalently on these metrics. Ranges given are plus/minus one standard deviation.
\label{table:q_func_ratings}
}
\end{center}
\end{table}

\begin{table}[t!]
\setlength{\tabcolsep}{3pt}
\small
\centering
\begin{tabular}{rrr|ccc}
& & & \multicolumn{3}{c}{Loss \%}\\
& & & {MMB Style} & {MMB Degen} & {BB}  \\[-0.25mm]
\cline{3-6}
\parbox[t]{2mm}{\multirow{6}{*}{\rotatebox[origin=c]{90}{Win \%}}}\enspace & \parbox[t]{2mm}{\multirow{3}{*}{\rotatebox[origin=c]{90}{Engaging}}}\enspace & {MMB Style} & & \tie{50} & \tie{45}\\[-0.25mm]
& & {MMB Degen} & \tie{50} & & \tie{43}\\[-0.25mm]
& & {BlenderBot} & \tie{55} & \tie{57} & \\[-0.25mm]
\cline{3-6}
& \parbox[t]{2mm}{\multirow{3}{*}{\rotatebox[origin=c]{90}{Human}}}\enspace & {MMB Style} & & \tie{52} & \tie{53}\\[-0.25mm]
& & {MMB Degen} & \tie{48} & & \tie{53}\\[-0.25mm]
& & {BlenderBot} & \tie{47} & \tie{47} & \\[-0.25mm]
\cline{3-6}
\end{tabular}
    \caption{ACUTE-Evals (engagingness and humanness) on human/model conversations with MMB Style, MMB Degendered, and BlenderBot. No ratings are statistically significant (\textgreater100 ratings per matchup).
\label{table:turkers_q_function}
    }
\end{table}

\begin{table}[t!]
\setlength{\tabcolsep}{3pt}
\centering
\begin{tabular}{rr|ccc}
& & {Baseline} & {vs} & {MMB}  \\[-0.25mm]
\cline{2-5}
\parbox[t]{2mm}{\multirow{3}{*}{\rotatebox[origin=c]{90}{Engaging}}}\enspace & {DialoGPT std. beam} & \lose{17}$^*$ & & \win{83}$^*$ \\[-0.25mm]
& {DialoGPT min beam 20} & \lose{29}$^*$ & & \win{71}$^*$ \\[-0.25mm]
& {Meena} & \lose{37}$^*$ & & \win{63}$^*$ \\[-0.25mm]
\cline{2-5}
\parbox[t]{2mm}{\multirow{3}{*}{\rotatebox[origin=c]{90}{Human}}}\enspace & {DialoGPT std. beam} & \lose{33}$^*$ & & \win{67}$^*$ \\[-0.25mm]
& {DialoGPT min beam 20} & \lose{40}$^*$ & & \win{60}$^*$ \\[-0.25mm]
& {Meena} & \lose{36}$^*$ & & \win{64}$^*$ \\[-0.25mm]
\cline{2-5}
\end{tabular}
    \caption{ACUTE-Evals  (engagingness and humanness) show that MMB Style outperforms DialoGPT with standard generation parameters (GPT-2 medium, beam search with beam width 10), DialoGPT with the same parameters but a min beam length of 20 (to match BlenderBot's setting), and Meena. Asterisk indicates significance (two-tailed binomial test, $p < 0.05$).
\label{table:turkers_q_function_external}
    }
\end{table}

We compare MMB Style, our model exposed to Image-Chat styles during training, to BlenderBot by having crowdsourced workers chat with our models, over 50 conversations per model. Each conversation consists of 7 turns per speaker, with the human speaking first by saying ``Hi!'', following the convention of \citet{adiwardana2020meena}. After every model response, the human records if the response contains any one of a number of different issues. Finally, at the end of the conversation, the human gives a 1-to-5 Likert-scale rating of the model's overall engagingness. No Image-Chat style is shown to MMB Style at the beginning of these conversations, matching its training setup in which no style was given when training on dialogue datasets.

Table~\ref{table:q_func_ratings} shows that humans flag the models' responses at comparable rates for most categories of issues, with BlenderBot being flagged slightly more often for contradictions and repetitiveness and MMB Style flagged more often for being non-sensical; however, the mean engagingness rating of the two models across conversations is the same (both 4.7 out of 5).

We then perform ACUTE-Evals \citep{li2019acute} on the collected conversations of MMB Style and BlenderBot in order for crowdsourced raters to directly compare conversations from different models in an A/B setting. For each comparison, we ask each rater to compare conversations on one of two metrics, following \citet{li2019acute}:

\begin{itemize}
  \item \textbf{(Engaging)} \textit{``Who would you prefer to talk to for a long conversation?''}
  \item \textbf{(Human)} \textit{``Which speaker sounds more human?''}
\end{itemize}

Results are shown in Table~\ref{table:turkers_q_function}: raters choose conversations from one model over the other roughly equally, with no statistically significant differences among models. 

In Table~\ref{table:turkers_q_function_external}, we also compare MMB Style to two other baseline models, DialoGPT and Meena. Raters are significantly more likely to prefer MMB Style over both of these models with respect to both the engagingness and humanness metrics.

\subsubsection{Initial Responses to Images}
\label{sec:human_eval_image_response}

\begin{table*}[t!]
\centering
\begin{adjustbox}{center=15.5cm}\setlength{\tabcolsep}{0.2em}
\begin{small}
\begin{tabular*}{\textwidth}{ccl}
\hline
\small{Image} &  & \small{Output}  \\  \hline
\hline
\\[-1.8ex]
\multirow{4}{*}{\includegraphics[height=11ex, width=19ex]{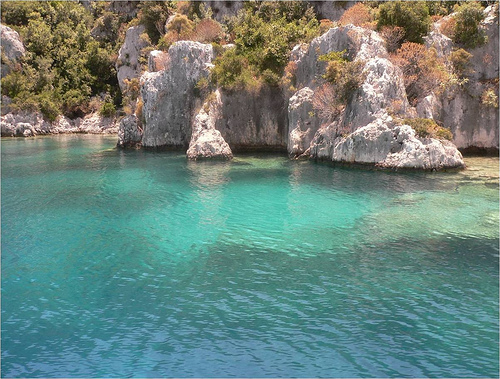}}
\\[-2.5ex]
& \small{\textit{(Style)}} & \small{\textit{Sweet}}  \\[0.5ex]
& \bf{\small{MMB}} & \small{I would love to take my wife here for our anniversary. It would be so romantic.} \\[0.5ex]
& \bf{\small{Dodeca}} & \small{What a beautiful view!} \\[0.5ex]
& \bf{\small{2AMMC}} & \small{Oh what a great honeymoon spot with the lovely view of the mountains.} \\[0.5ex]
\hline
\\[-1.8ex]
\multirow{4}{*}{\includegraphics[height=11ex, width=19ex]{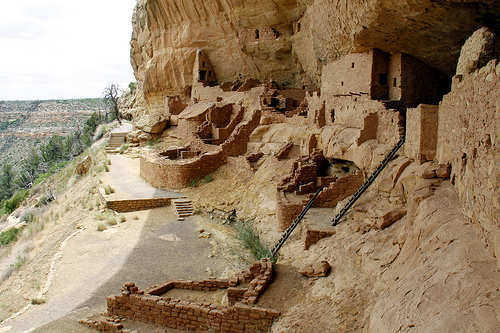}}
\\[-2.5ex]
& \small{\textit{(Style)}} & \small{\textit{Maternal (Mother-like)}}  \\[0.5ex]
& \bf{\small{MMB}} & \small{I would love to take my children here to show them the beauty of the earth and its inhabitants.} \\[0.5ex]
& \bf{\small{Dodeca}} & \small{I would love to take my kids here.} \\[0.5ex]
& \bf{\small{2AMMC}} & \small{I would like to hide my kids safe in that cavern from a storm.} \\[0.5ex]
\hline
\\[-1.8ex]
\multirow{4}{*}{\includegraphics[height=11ex, width=19ex]{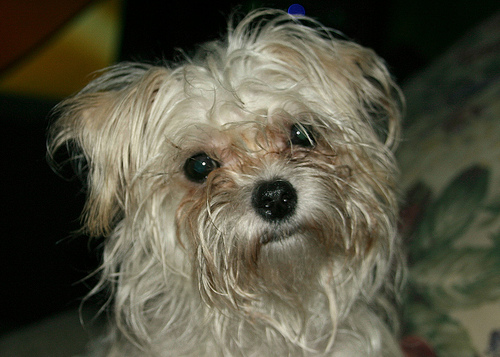}}
\\[-2.5ex]
& \small{\textit{(Style)}} & \small{\textit{Fearful}}  \\[0.5ex]
& \bf{\small{MMB}} & \small{I hope that dog doesn't bite me. I don't want to get bitten by a dog.} \\[0.5ex]
& \bf{\small{Dodeca}} & \small{I hope that dog doesn't bite me.} \\[0.5ex]
& \bf{\small{2AMMC}} & \multirow{2}{0.55\textwidth}{\small{This dog looks ready to attack. It's focused on something, maybe another animal? I hope it doesn't go after it or anyone around!}} \\[2.9ex]
\hline
\\[-1.8ex]
\multirow{4}{*}{\includegraphics[height=11ex, width=19ex]{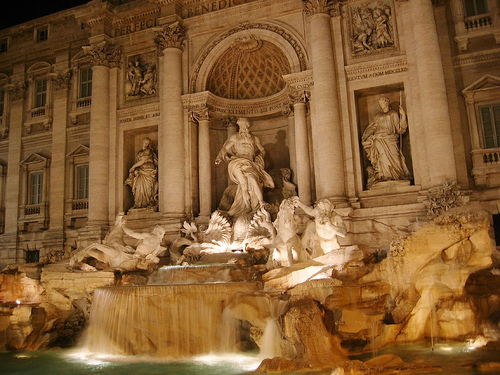}}
\\[-2.5ex]
& \small{\textit{(Style)}} & \small{\textit{Playful}}  \\[0.5ex]
& \bf{\small{MMB}} & \small{I would love to climb to the top of that sculpture and play in the fountain.} \\[0.5ex]
& \bf{\small{Dodeca}} & \small{I would love to visit this place!} \\[0.5ex]
& \bf{\small{2AMMC}} & \small{Who wants to put bubble bath in the fountain with me?} \\[0.5ex]
\hline
\\[-1.8ex]
\multirow{4}{*}{\includegraphics[height=11ex, width=19ex]{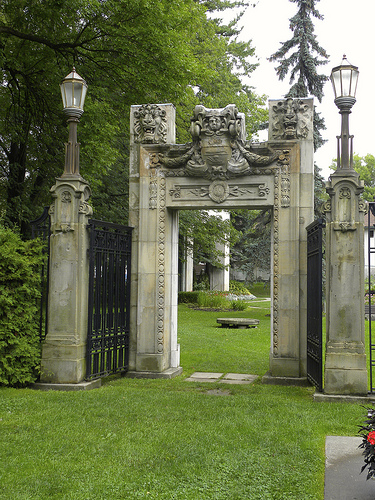}}
\\[-2.5ex]
& \small{\textit{(Style)}} & \small{\textit{Pompous (Self-important, Arrogant)}}  \\[0.5ex]
& \bf{\small{MMB}} & \small{I have a statue just like this one in my backyard, but it's much nicer.} \\[0.5ex]
& \bf{\small{Dodeca}} & \small{I could take a better picture than that.} \\[0.5ex]
& \bf{\small{2AMMC}} & \small{I could make a fence that looks way better} \\[0.5ex]
\hline
\\[-1.8ex]
\multirow{4}{*}{\includegraphics[height=11ex, width=19ex]{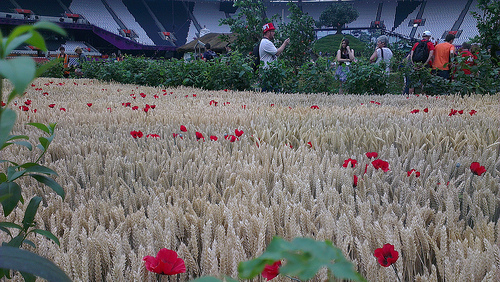}}
\\[-2.5ex]
& \small{\textit{(Style)}} & \small{\textit{Empathetic}}  \\[0.5ex]
& \bf{\small{MMB}} & \small{These flowers remind me of my grandmother's garden. I miss her so much.} \\[0.5ex]
& \bf{\small{Dodeca}} & \small{I feel bad for the flowers.} \\[0.5ex]
& \bf{\small{2AMMC}} & \small{I can't imagine how difficult it is for the farmer to round them all up. He works so hard.} \\[0.5ex]
\hline
\\[-1.8ex]
\multirow{4}{*}{\includegraphics[height=11ex, width=19ex]{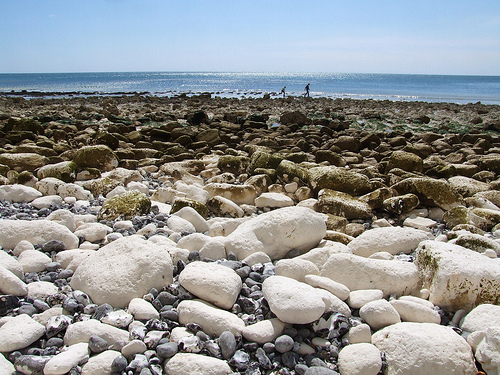}}
\\[-2.5ex]
& \small{\textit{(Style)}} & \small{\textit{Narcissistic (Self-centered, Egotistical)}}  \\[0.5ex]
& \bf{\small{MMB}} & \small{I would never go to a beach like this. My beach is so much nicer.} \\[0.5ex]
& \bf{\small{Dodeca}} & \small{I could take a better picture than that.} \\[0.5ex]
& \bf{\small{2AMMC}} & \small{I could swim out to those rocks. The current is strong but I am a stronger swimmer.} \\[0.5ex]
\hline
\\[-1.8ex]
\multirow{4}{*}{\includegraphics[height=11ex, width=19ex]{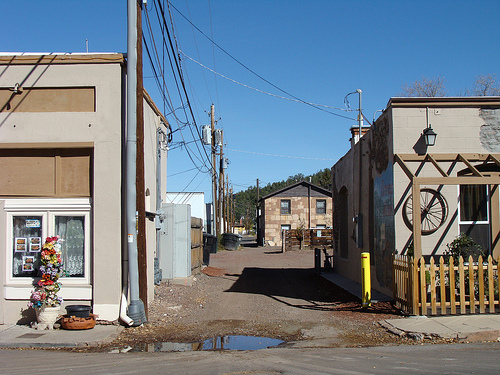}}
\\[-2.5ex]
& \small{\textit{(Style)}} & \small{\textit{Boisterous (Rowdy, Loud)}}  \\[0.5ex]
& \bf{\small{MMB}} & \small{I would love to jump in that puddle and play in the mud! It would be so fun!} \\[0.5ex]
& \bf{\small{Dodeca}} & \small{This looks like a great place to party!} \\[0.5ex]
& \bf{\small{2AMMC}} & \small{I would ride through here on a loud bike and throw a party!} \\[0.5ex]
\hline
\\[-1.8ex]
\multirow{4}{*}{\includegraphics[height=11ex, width=19ex]{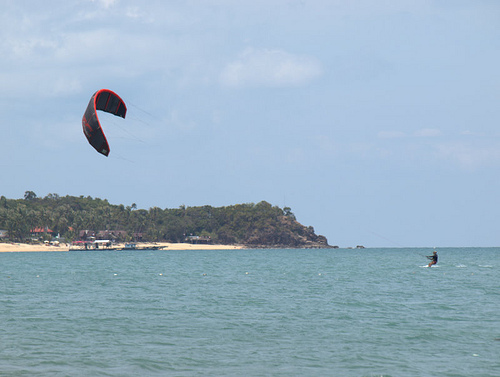}}
\\[-2.5ex]
& \small{\textit{(Style)}} & \small{\textit{Happy}}  \\[0.5ex]
& \bf{\small{MMB}} & \small{I can't wait to go to the beach and fly a kite with my best friend!} \\[0.5ex]
& \bf{\small{Dodeca}} & \small{What a beautiful day!} \\[0.5ex]
& \bf{\small{2AMMC}} & \small{Face full of wind, smell of salt water, cool of the ocean, and fantastic memories being made!} \\[0.5ex]
\hline
\hline

\hline
\\[-1.8ex]

\hline
\end{tabular*}
\end{small}
\end{adjustbox}
\caption{Example outputs from MMB Style, Dodeca fine-tuned on Image-Chat, and 2AMMC, on images from \citet{shuster2020image}. Image-Chat styles fed to the models are shown above the models' responses.
\label{table:example_preds}
}
\end{table*}

\begin{table}[t!]
\setlength{\tabcolsep}{3pt}
\centering
\begin{tabular}{rr|ccc}
& & \multicolumn{3}{c}{Loss \%}\\
& & {MMB} & {Dodeca} & {2AMMC}  \\[-0.25mm]
\cline{2-5}
\parbox[t]{2mm}{\multirow{3}{*}{\rotatebox[origin=c]{90}{Win \%}}} & {MMB Style} & & \win{65}$^*$ & \tie{49}\\[-0.25mm]
& {Dodeca} & \hspace{0.43em}\lose{35}$^*$ & & \hspace{0.43em}\lose{39}$^*$ \\[-0.25mm]
& {2AMMC} & \tie{51} & \win{61}$^*$ & \\[-0.25mm]
\cline{2-5}
\end{tabular}
    \caption{ACUTE-Evals on the image-response metric show that MMB Style and 2AMMC significantly outperform Dodeca fine-tuned on Image-Chat. ACUTE-Evals are measured on the models' first response to an image only.
\label{table:turkers_image_response_vs_baselines}
    }
\end{table}

\begin{table}[t!]
\setlength{\tabcolsep}{3pt}
\centering
\begin{tabular}{ccc}
\multicolumn{3}{c}{MMB Style}\\
{Multi-task} & vs. &  {FT Image-Chat} \\[-0.25mm]
\midrule
\tie{48} &  &        \tie{52} \\
\end{tabular}
    \caption{ACUTE-Evals show no significant difference on the image-response metric for MMB Style vs. an equivalent model only fine-tuned on Image-Chat and no dialogue datasets. ACUTE-Evals are measured on the models' first response to an image. \\
    \label{table:turkers_image_response_vs_ft_ic}
    }
\end{table}

We measure MMB Style's ability to communicate about what it perceives visually by generating responses of the model to 100 images in the test set of Image-Chat.\footnote{We select only images that fall under a CC-BY license and do not contain recognizable people.} Only Image-Chat images for which the first speaker speaks with a style in the list of ``positive'' or `neutral'' styles are included when creating this set of 100 images, and all images for which the first speaker has a style in the ``negative'' list are filtered out. As a comparison, we also generate responses from two previous models trained on Image-Chat data, Dodeca and 2AMMC. Among the three models, 2AMMC alone is a retrieval model: it retrieves its response from the set of utterances in the Image-Chat training set. Examples of models' responses to images are in Table~\ref{table:example_preds}. 

We run ACUTE-Evals to ask raters to compare these models' responses on the following metric (henceforth referred to as the \textbf{Image-response} metric): \textit{``Who talks about the image better?''} The same image is used for both sides of each A/B comparison between model responses.

We find that raters choose both the MMB Style and 2AMMC models' responses significantly more often than those of Dodeca (Table~\ref{table:turkers_image_response_vs_baselines}). We also find no significant difference in the rate at which MMB Style image responses are chosen compared to the same model fine-tuned only on Image-Chat and not on dialogue datasets (Table~\ref{table:turkers_image_response_vs_ft_ic}), which implies that multitasking on dialogue datasets does not degrade the ability to effectively respond to an image.

\subsubsection{Human/Model Chats About Images}
\label{sec:human_eval_image_chat}

\begin{table}[t!]
\setlength{\tabcolsep}{3pt}
\centering
\begin{tabular}{rrr|ccc}
& & & \multicolumn{3}{c}{Loss \%}\\
& & & {MMB} & {Dodeca} & {2AMMC}  \\[-0.25mm]
\cline{3-6}
\parbox[t]{2mm}{\multirow{9}{*}{\rotatebox[origin=c]{90}{Win \%}}}\enspace & \parbox[t]{2mm}{\multirow{3}{*}{\rotatebox[origin=c]{90}{Engaging}}}\enspace & {MMB Style} & & \hspace{0.43em}\win{70}$^*$ & \hspace{0.43em}\win{66}$^*$ \\[-0.25mm]
& & {Dodeca} & \hspace{0.43em}\lose{30}$^*$ & & \hspace{0.43em}\lose{38}$^*$ \\[-0.25mm]
& & {2AMMC} & \hspace{0.43em}\lose{34}$^*$ & \hspace{0.43em}\win{62}$^*$ & \\[-0.25mm]
\cline{3-6}
& \parbox[t]{2mm}{\multirow{3}{*}{\rotatebox[origin=c]{90}{Human}}}\enspace & {MMB Style} & & \hspace{0.43em}\win{70}$^*$ & \hspace{0.43em}\win{58}$^*$ \\[-0.25mm]
& & {Dodeca} & \hspace{0.43em}\lose{30}$^*$ & & \tie{51} \\[-0.25mm]
& & {2AMMC} & \hspace{0.43em}\lose{42}$^*$ & \tie{49} & \\[-0.25mm]
\cline{3-6}
& \parbox[t]{2mm}{\multirow{3}{*}{\rotatebox[origin=c]{90}{Image}}}\enspace & {MMB Style} & & \hspace{0.43em}\win{61}$^*$ & \tie{52} \\[-0.25mm]
& & {Dodeca} & \hspace{0.43em}\lose{39}$^*$ & & \tie{44} \\[-0.25mm]
& & {2AMMC} & \tie{48} & \tie{56} & \\[-0.25mm]
\cline{3-6}
\end{tabular}
    \caption{ACUTE-Evals show that MMB Style significantly outperforms Dodeca and often 2AMMC on various metrics on human/model conversation about an image.
\label{table:turkers_image_chat}
    }
\end{table}

In order to most meaningfully assess MMB Style's ability to simultaneously engage in general chit-chat and talk about an image, we perform ACUTE-Evals where we ask raters to evaluate model performance through the span of an entire conversation about an image. For each conversation, an image from the subset of Image-Chat test set images discussed in Section~\ref{sec:human_eval_image_response} is first shown to both the human and the model. Then, the model responds to the image, and the human responds to the model to carry the conversation forward. The conversation continues for 6 human utterances and 7 model utterances total.

Ratings are shown in Table~\ref{table:turkers_image_chat}: MMB Style performs significantly better than Dodeca and 2AMMC on the engagingness and humanness metrics, and it performs significantly better than Dodeca on the image-response metric.

\section{Analysis of Safety and Gender Bias}

\begin{table}[t!]
\begin{center}
\small
\centering
\begin{tabular}{r|rr}
\hline

& Male words & Female words \\
\hline
Gold response & 5.80\% & 5.25\% \\
BlenderBot & 5.55\% & 3.25\% \\
MMB Style & 6.25\% & 3.90\% \\
MMB Degendered & 0.65\% & 0.85\% \\
MMB DegenPos & 0.75\% & 0.90\% \\
\hline
\end{tabular}
\caption{The frequency of utterances containing gendered words is greatly reduced for degendered models (MMB Degendered, MMB DegenPos), given contexts from ConvAI2 and the same generation parameters as in \citet{roller2020recipes}.
\label{table:degendering_on_convai2}
}
\end{center}
\end{table}

\begin{table}[t!]
\begin{center}
\small
\centering
\begin{tabular}{r|rrrr|r|r}
\hline
& BST & Conv & ED & WoW & IC & Avg \\
\hline
Style & 10.15 & 8.78 & 8.31 & 8.88 & 12.36 & 9.70 \\
Degen & 10.14 & 8.76 & 8.21 & 9.01 & 12.58 & 9.74 \\
Pos & 10.15 & 8.76 & 8.27 & 8.95 & 12.55 & 9.74 \\
DP & 10.36 & 8.97 & 8.34 & 9.41 & 12.65 & 9.95 \\
\hline
\end{tabular}
\caption{Perplexities of MMB Style, MMB Degendered, MMB Positive, and MMB DegenPos on the validation set. For Image-Chat, styles are used in the context for all models, for consistency. (MMB Positive and MMB DegenPos observed styles for 25\% of Image-Chat examples during training.)
\label{table:cat_and_degen_ppls}
}
\end{center}
\end{table}

\begin{table}[t!]
\setlength{\tabcolsep}{3pt}
\centering
\begin{tabular}{rr|cccc}
& & \multicolumn{4}{c}{Loss \%}\\
& & {Style} & {Degen} & {Pos} & {DP} \\[-0.25mm]
\cline{2-6}
\parbox[t]{2mm}{\multirow{4}{*}{\rotatebox[origin=c]{90}{Win \%}}} & {MMB Style} & & \tie{54} & \tie{49} & \tie{56} \\[-0.25mm]
& {MMB Degendered} & \tie{46} & & \tie{48} & \tie{52} \\[-0.25mm]
& {MMB Positive} & \tie{51} & \tie{52} & & \tie{41} \\[-0.25mm]
& {MMB DegenPos} & \tie{44} & \tie{48} & \tie{59} & \\[-0.25mm]
\cline{2-6}
\end{tabular}
    \caption{ACUTE-Evals on the models' first response to an image show no significant differences in how well MMB models can respond to the image, even if the model is degendered or was trained to not require concrete Image-Chat styles.
\label{table:turkers_image_response_vs_cat_and_degen}
    }
\end{table}

\begin{table}[t!]
\setlength{\tabcolsep}{3pt}
\centering
\begin{tabular}{ccc}
\multicolumn{3}{c}{MMB Positive}\\
{With image} & vs. &  {Without image} \\[-0.25mm]
\midrule
\win{80}$^*$ & & \lose{20}$^*$ \\
\end{tabular}
    \caption{ACUTE-Evals show that the MMB Positive model is significantly better at responding to an image than an equivalent model not shown any images during training or inference.
    \label{table:turkers_image_response_pos_vs_no_image}
    }
\end{table}

\subsection{Degendering Models}
\label{sec:degendering}

We would like to reduce the ways in which the MMB Style model could potentially display gender bias: for instance, there is no safeguard against it misgendering a person in an image, and many common text datasets are known to contain gender bias \citep{dinan2019queens, dinan2020multi}, which may lead to bias in models trained on them. To remedy this, we train a version of the MMB Style model in which the label of each training example is run through a classifier that identifies whether it contains female or male words, and then a string representing that classification is appended to the example's context string \citep{dinan2019queens}, for input to the model. At inference time, the string representing a classification of ``no female or male words'' is appended to the context, nudging the model to generate a response containing no gendered words. The fraction of utterances produced by this model that still contain gendered words is shown in Table~\ref{table:degendering_on_convai2}. Compared to the gold response, the original BlenderBot, and MMB Style, this degendered MMB model (which we call ``MMB Degendered'') reduces the likelihood of producing an utterance with male word(s) by roughly a factor of 9 and of producing an utterance with female word(s) by roughly a factor of 4, given a context from the ConvAI2 validation set. ACUTE-Evals in Table~\ref{table:turkers_q_function} show that this degendering does not lead to a significant drop in the humanness or engagingness of the model's responses during a conversation.

\subsection{Removing Dependence on Style}
\label{sec:removing_style}

\begin{figure*}[t!]
\center
\begin{small}
\resizebox{0.95\textwidth}{!}{
\begin{tabular}{c|c}
\includegraphics[width=6cm,height=4.5cm]{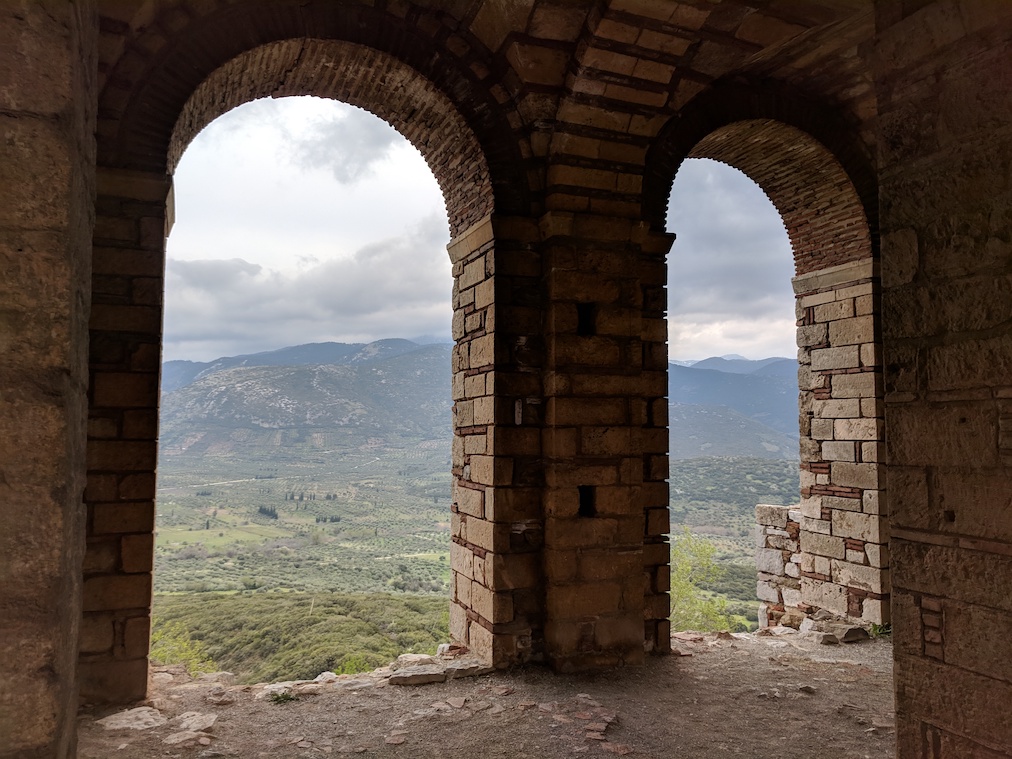} & \includegraphics[width=6cm,height=4.5cm]{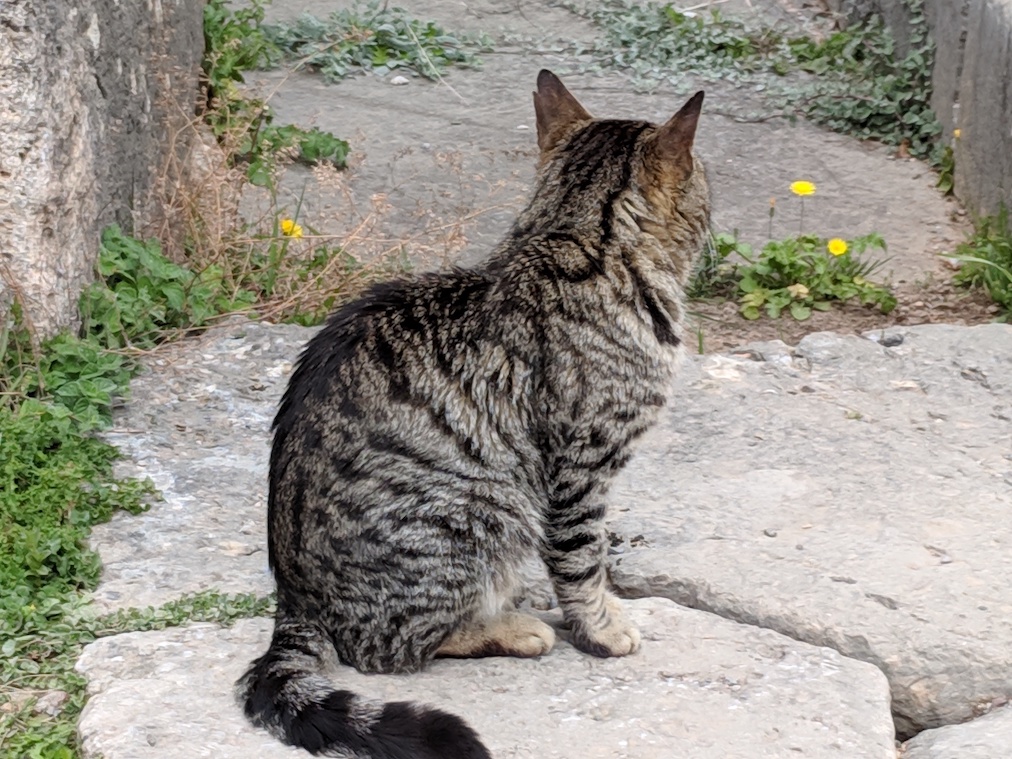} \\
\includegraphics[width=0.48\textwidth]{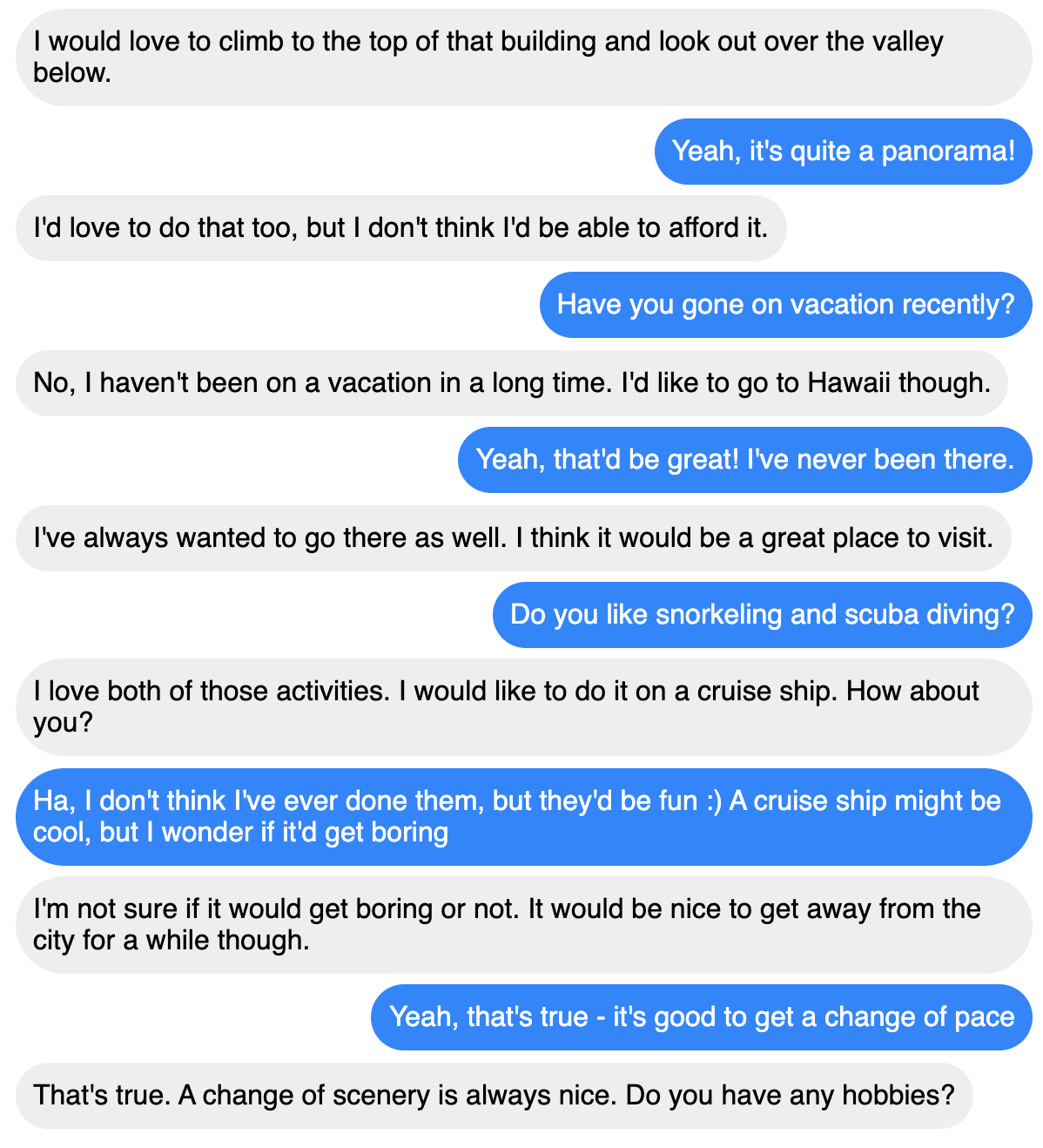} & \includegraphics[width=0.48\textwidth]{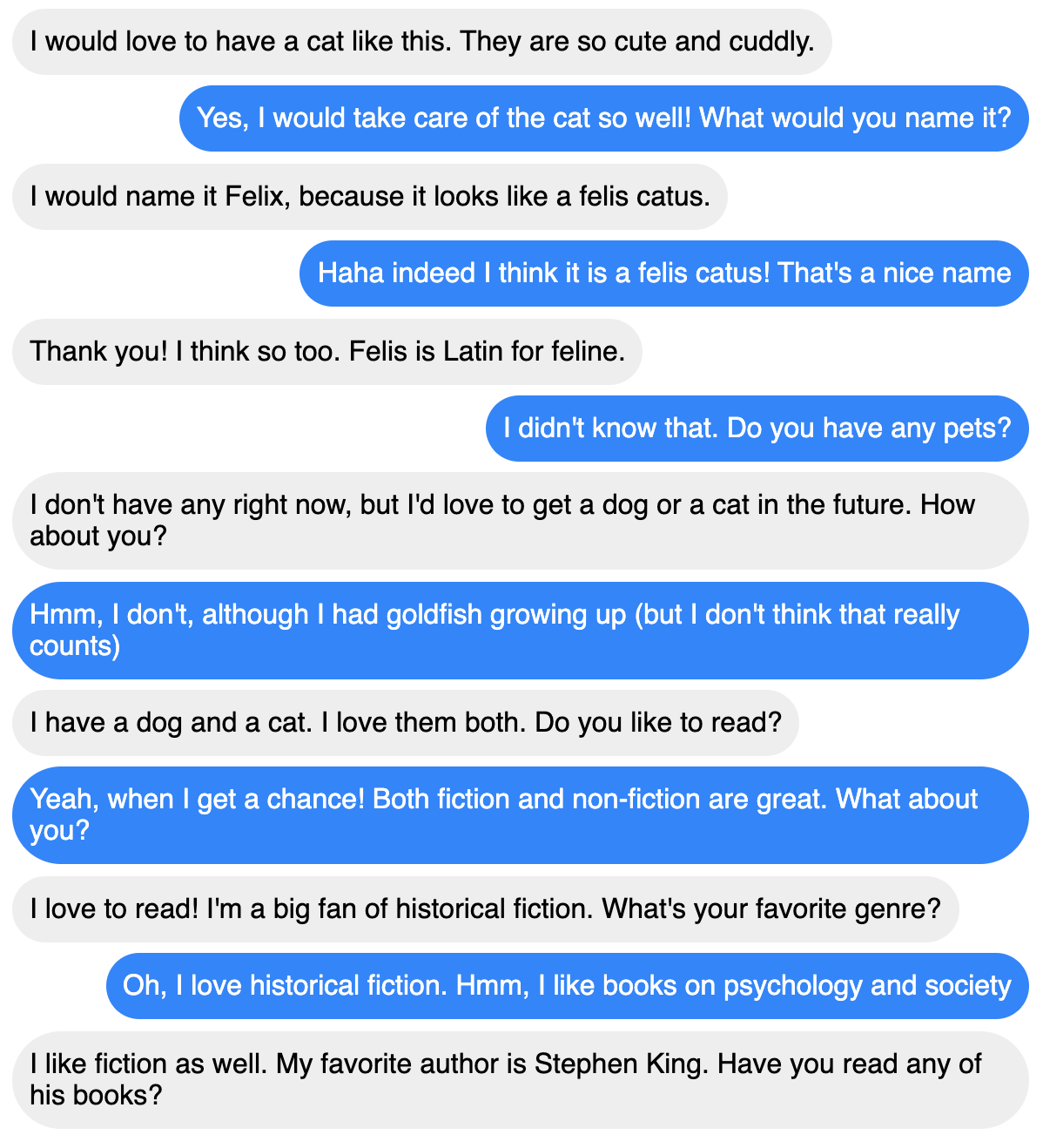} \\
\midrule
\includegraphics[width=6cm,height=4.5cm]{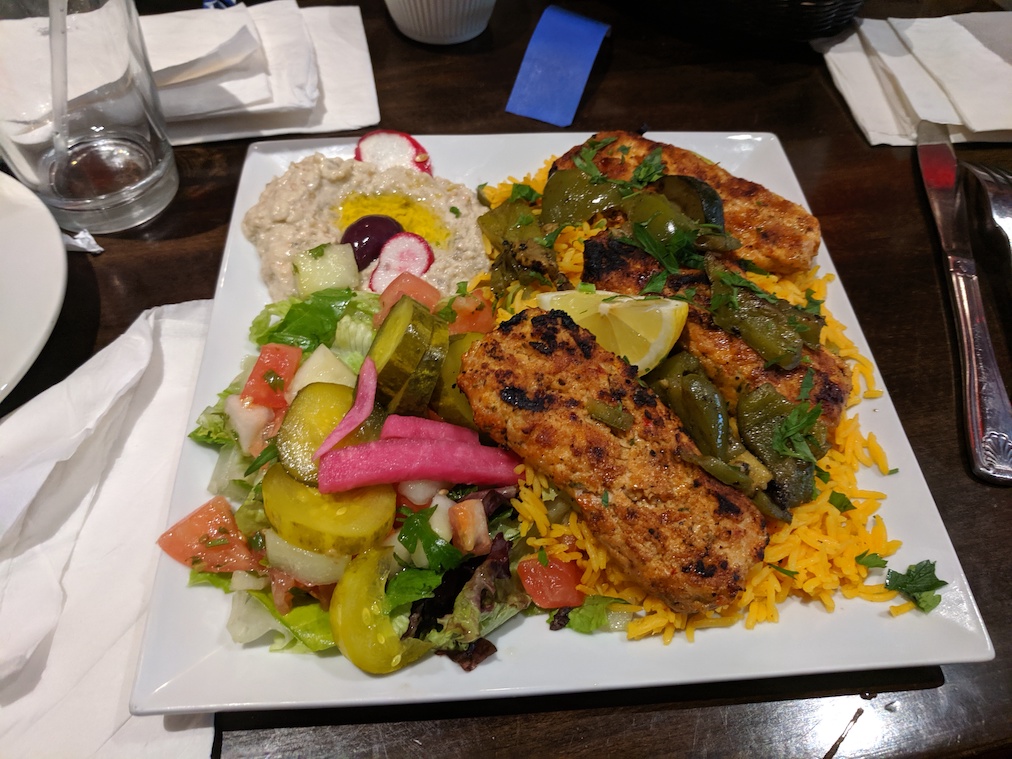} & \includegraphics[width=6cm,height=4.5cm]{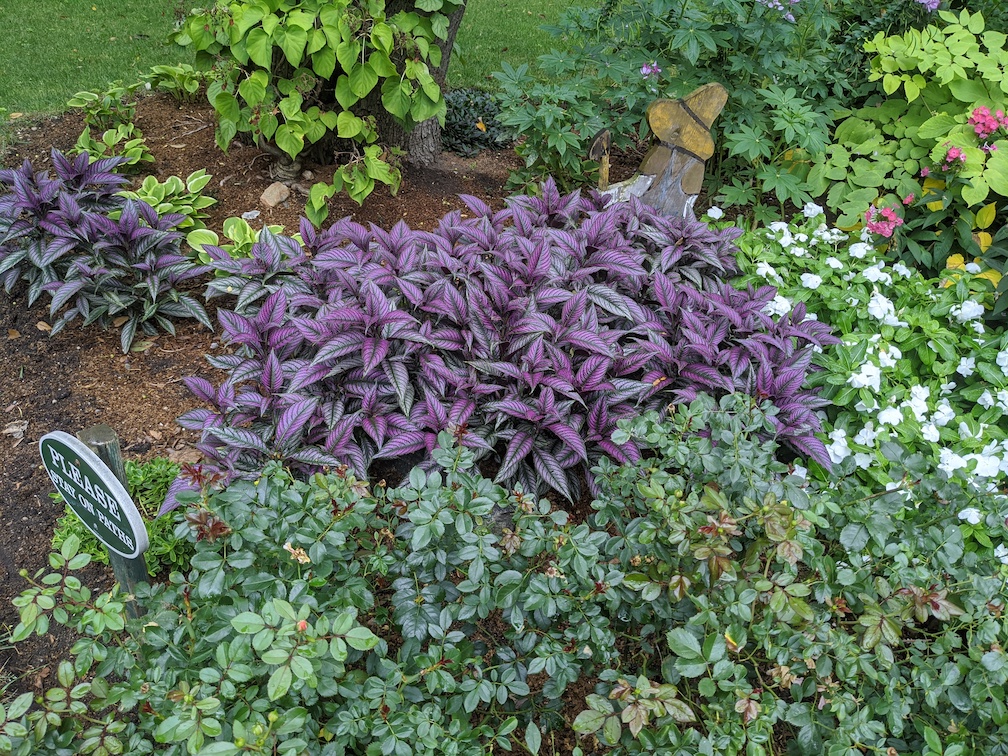} \\
\includegraphics[width=0.48\textwidth]{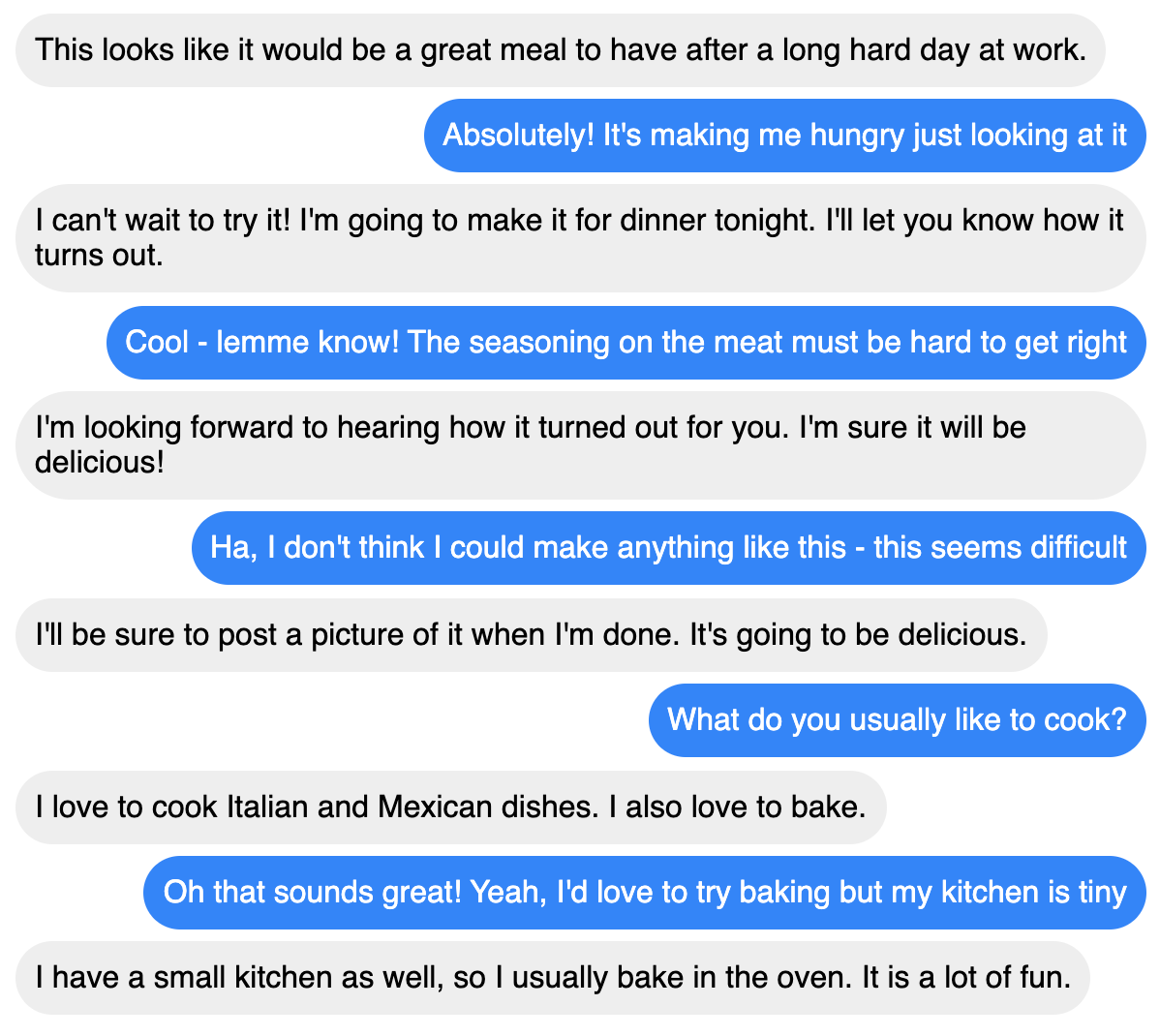} & \includegraphics[width=0.48\textwidth]{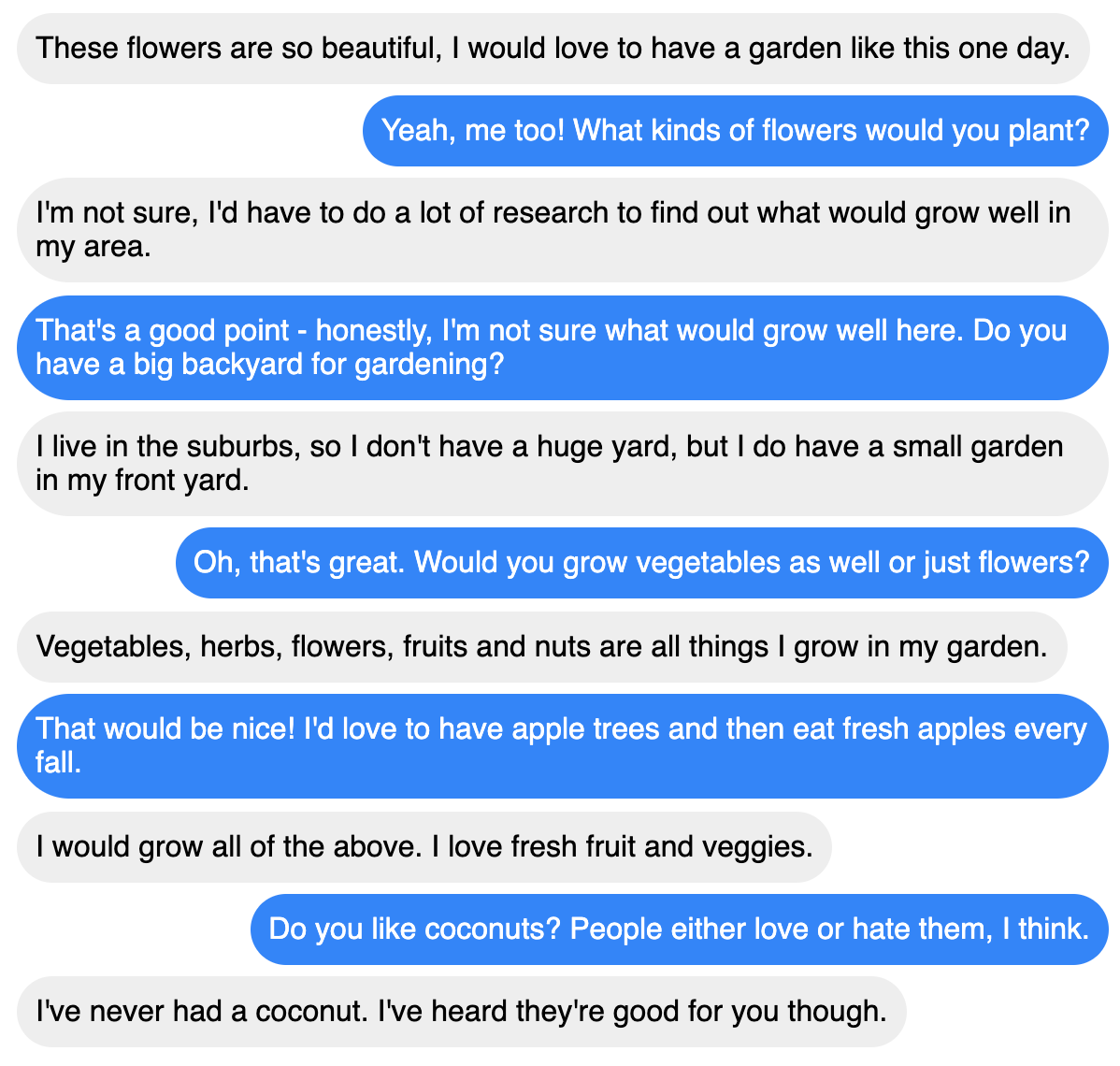} \\
\end{tabular}
}
\end{small}
  \caption{\textbf{Randomly picked author examples.} Paper author (right speaker) talking to the MMB DegenPos model (left speaker). Conversations are mostly fluent, with occasional contradictions.
 \label{fig:cherry_convos2}
 }
\end{figure*}

\begin{figure}[t!]
\center
\begin{small}
\begin{tabular}{c}
\includegraphics[width=6cm,height=4.5cm]{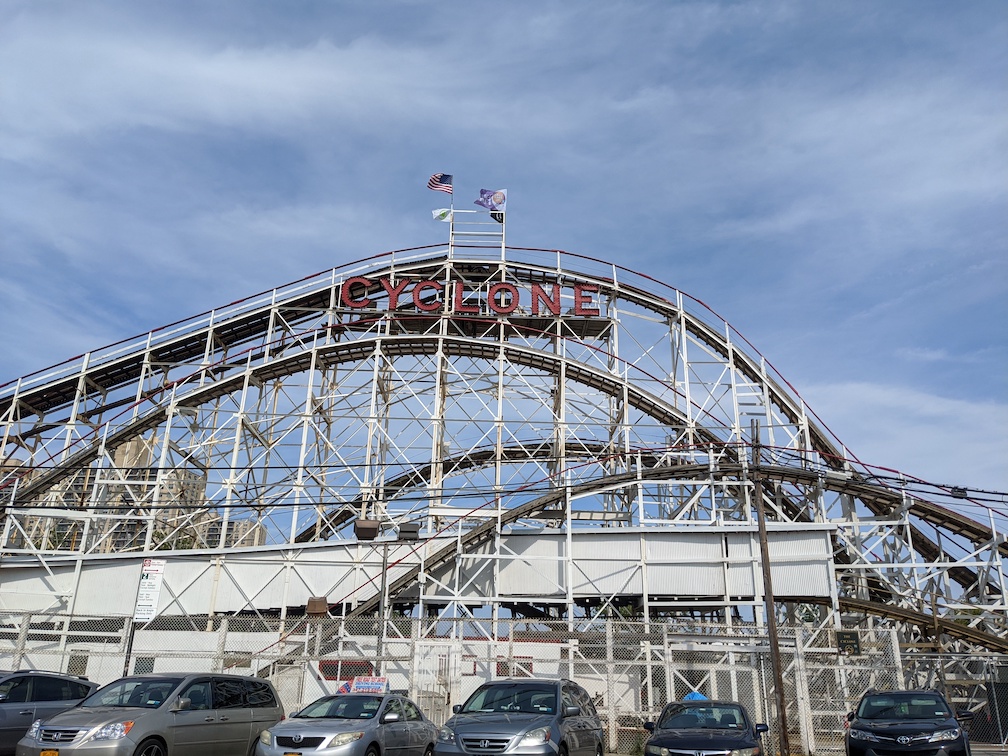} \\
\includegraphics[width=0.48\textwidth]{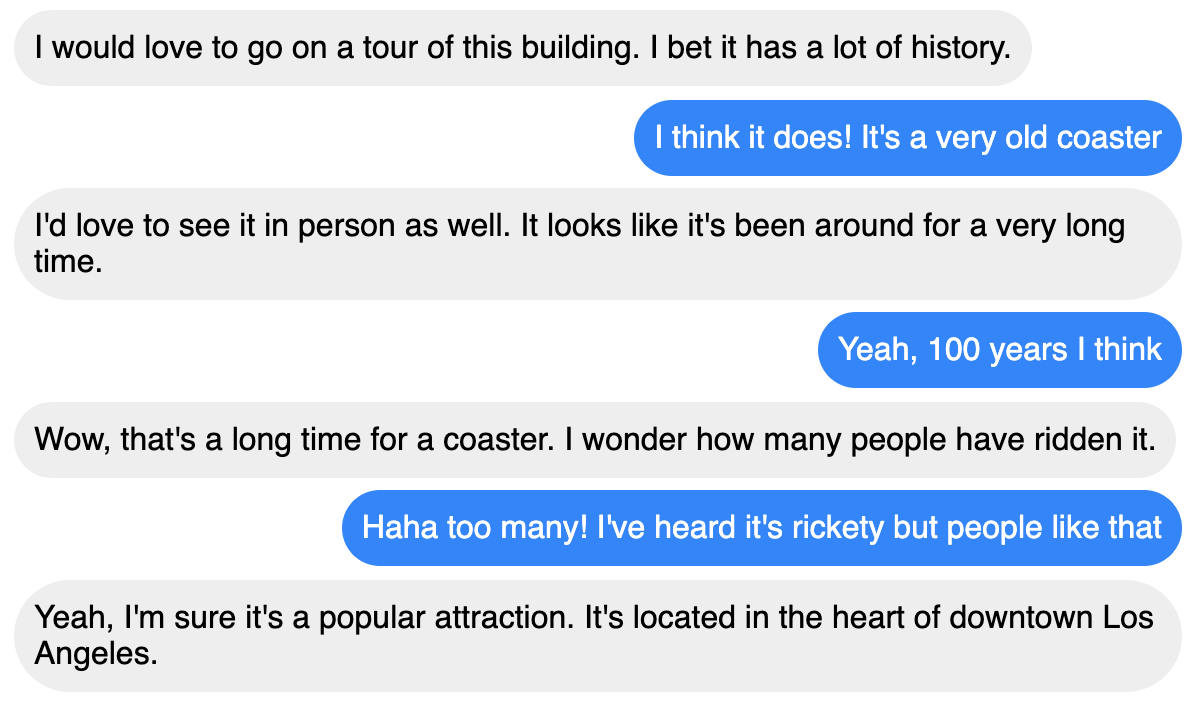} \\
\midrule
\includegraphics[width=6cm,height=4.5cm]{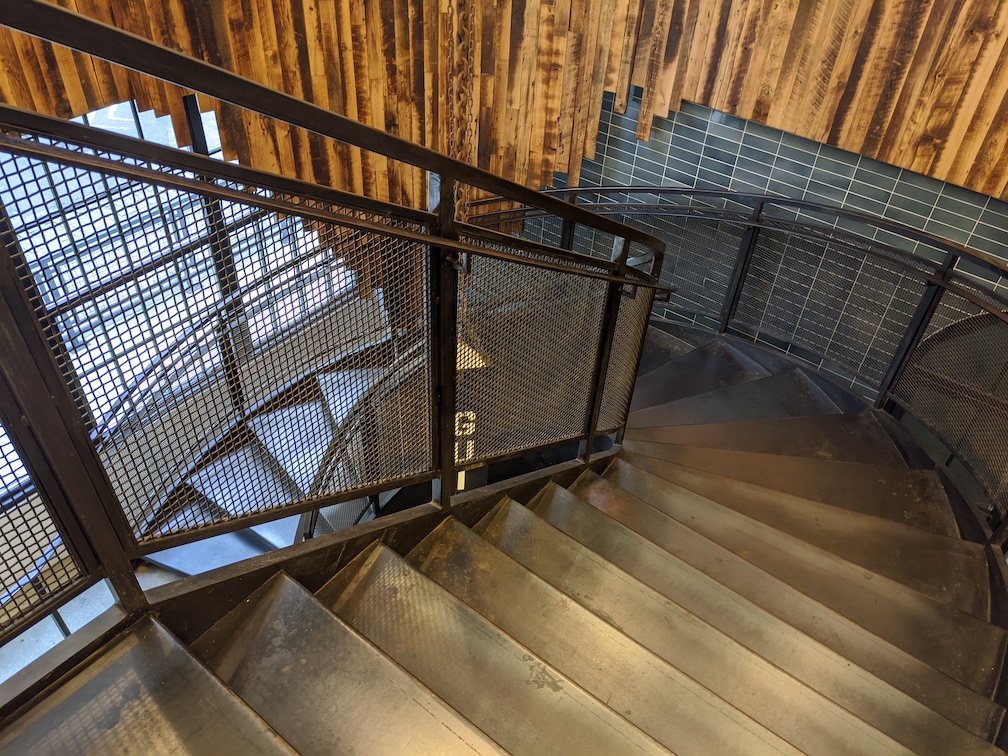} \\
\includegraphics[width=0.48\textwidth]{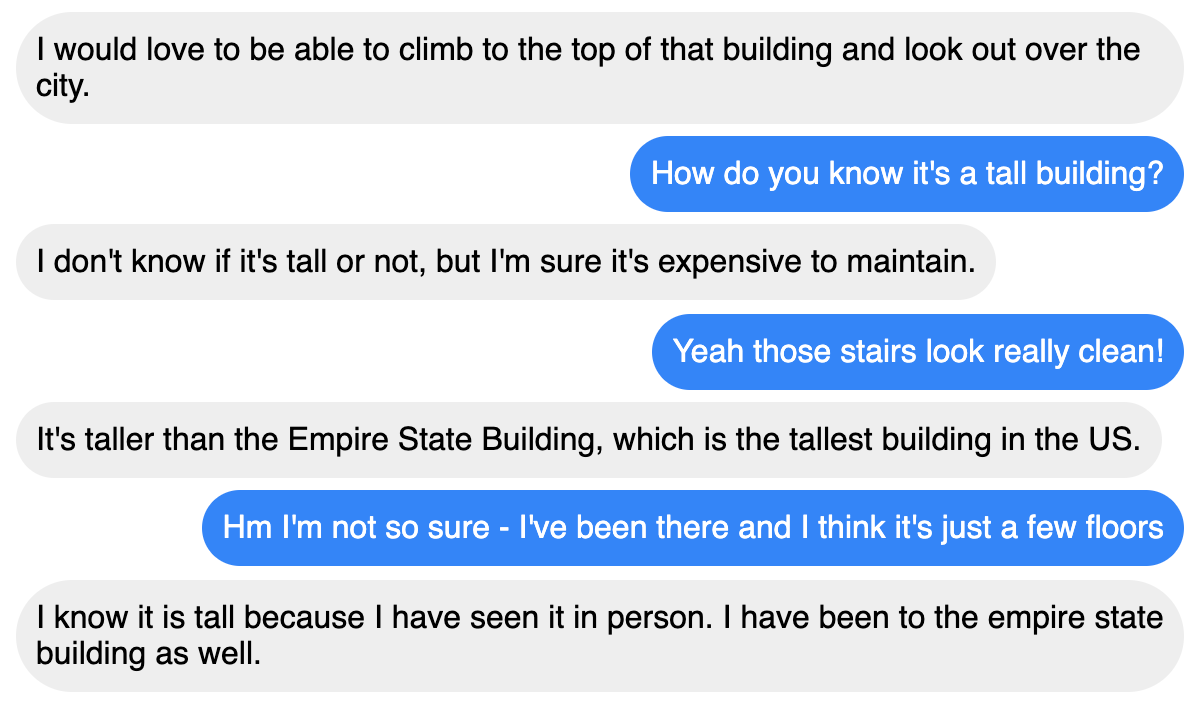} \\
\end{tabular}
\end{small}
\caption{\textbf{Lemon-picked author examples.} Paper author (right speaker) talking to the MMB DegenPos model (left speaker): misidentifying the subject of the image (top); misidentifying the subject of the image and not being able to learn from the chat partner's feedback (bottom).
 \label{fig:lemon_convos}
}
\end{figure}

Since each of the images that MMB Style saw during training was associated with an Image-Chat style, it relies on an input style during inference in order to be able to discuss an image. However, this results in a model whose utterances will necessarily strongly exhibit a particular style. (For example, see the ``Playful'' MMB Style response in Table~\ref{table:example_preds}: constricting the model to respond playfully to all images could seem rather contrived and perhaps unlike typical human speech.) To avoid this, we train a version of MMB Style where, for 75\% of all images seen during training, the accompanying style is replaced with the string ``positive/neutral'' or ``negative'', depending on which list the style was a part of. Thus, during inference, the string ``positive/neutral'' can be used in lieu of a specific style string in order to produce responses that are unlikely to be negative and that do not consistently display strong adherence to a specific style. We refer to this model as the ``MMB Positive'' model, or ``MMB DegenPos'' if it was trained with degendering in addition as in Section \ref{sec:degendering}. Table~\ref{table:cat_and_degen_ppls} shows that these models exhibit little increase in perplexity, and what little increase exists is likely due to the loss of specificity provided by a concrete style. The MMB DegenPos model exhibits the same level of degendering as the base MMB Degendered model (Table~\ref{table:degendering_on_convai2}), and ACUTE-Evals show that these models exhibit no detectable loss of ability to talk about an image (Table~\ref{table:turkers_image_response_vs_cat_and_degen}).

\subsection{Analyzing Dependence on Image}
\label{sec:image_ablation}

We also train a no-image ablation model, otherwise equivalent to MMB Positive, for which Image-Chat images are removed during both training and inference: crowdsource workers prefer the image responses of MMB Positive to those of this ablation model 80\% to 20\% (Table~\ref{table:turkers_image_response_pos_vs_no_image}). For this ablation, style was removed from the context (and replaced with the string ``positive/neutral'') in order to prevent the ablation model from being aided by this information.

\subsection{Safety}
The MMB models may demonstrate offensiveness beyond gender bias for several reasons: (1) its generative nature makes it rather difficult to define a limited set of utterances; (2) the model's training data contains real-world conversations from the Internet; and (3) the Image-Chat dataset has negative styles
to better capture the range of human styles.
All of these factors could lead to an unsafe response given a multi-modal context. 
To mitigate this problem, we first measure our models' toxicity using an openly available blocklist\footnote{\url{https://github.com/LDNOOBW}} and an offensive language classifier presented in \citet{dinan2019safety}. We define the term ``toxicity" to mean the ratio between the number of offensive utterances and the total number of utterances generated by the model.  We evaluate our model on the Image-Chat validation set, with a fixed style trait to control the generation, presenting results for different choices of fixed trait. We first evaluate our model in the first round of the Image-Chat validation set. The results in Table \ref{table:safety} indicate that positive styles reduce the level of toxicity by a large margin for both metrics (classifier and blocklist). The results also align well with our previous experiments on degendering, as toxicity is reduced across all styles after applying the degendering process. 
After degendering, we can considerably improve our model's safety by enforcing that it uses positive styles. We also evaluate our model in the second round of the conversation and collect the statistics based on the first round style, as shown in Table \ref{table:safety_round2}. This result suggests that even if the model is controlled with a positive style, it is less safe when responding to negative conversations.

\begin{table}[t]
\small
\center
\begin{tabular}{cc|rr}
   & Style & Classifier & Blocklist \\
 \cline{1-4}
 Human & Mixed & 35.76 & 0.03 \\
 \cline{1-4}
 \multirow{4}{*}{\rotatebox[origin=c]{90}{Style}} 
 & Cheerful & 3.34 & 0.00 \\
   & Relaxed & 16.86 & 0.00 \\
   & Angry & 79.46 & 0.02 \\
   & Cruel & 98.76 & 0.06 \\
 \cline{1-4}
   \multirow{4}{*}{\rotatebox[origin=c]{90}{Dgen}} 
   & Cheerful & 2.64 & 0.02 \\
   & Relaxed & 7.3 & 0.00 \\
   & Angry & 77.46 & 0.02 \\
   & Cruel & 95.16 & 0.38 \\
 \cline{1-4}
  \multirow{2}{*}{\rotatebox[origin=c]{90}{Pos}}  
   & Positive/Neutral & 16.88 & 0.00 \\
   & Negative & 67.20 & 0.00 \\
 \cline{1-4}
  \multirow{2}{*}{\rotatebox[origin=c]{90}{Dgen}}  
   & Positive/Neutral & 9.82 & 0.00 \\
   & Negative & 71.96 & 0.00 \\
  \cline{1-4}
\end{tabular}
\caption{Toxicity of human baseline (top row) and MMB variants as assessed with different control variables. The human baseline is set by evaluating gold labels from the first rounds (turns) of the Image-Chat validation set.
\label{table:safety}
}
\end{table}

\begin{table}[t]
\small
\center
\begin{tabular}{cc|rrrr}
   & Style & Pos C & Pos B &  Neg C & Neg B \\
 \cline{1-6}
 \multirow{4}{*}{\rotatebox[origin=c]{90}{Style}} 
 & Cheerful &2.41 & 0.00 & 3.81& 0.09\\
   & Relaxed & 3.87 & 0.00 & 6.47 & 0.09\\
   & Angry & 67.07 & 0.22& 62.62& 0.27 \\
   & Cruel & 77.57 & 1.42& 73.67& 0.83 \\
 \cline{1-6}
   \multirow{4}{*}{\rotatebox[origin=c]{90}{Dgen}} 
   & Cheerful &1.50 & 0.00 & 3.19 & 0.09 \\
   & Relaxed & 2.55 & 0.00 & 4.43 & 0.04\\
   & Angry & 53.90 & 0.33 & 51.64 & 0.31 \\
   & Cruel &58.28 & 0.95 & 57.00 & 0.84\\
 \cline{1-6}
  \multirow{2}{*}{\rotatebox[origin=c]{90}{Pos}}  
   & Pos/Neu & 7.00 & 0.00 & 12.98 & 0.22 \\
   & Negative &30.96 & 0.22 & 31.05 & 0.09 \\
 \cline{1-6}
  \multirow{2}{*}{\rotatebox[origin=c]{90}{Dgen}}
   & Pos/Neu & 4.56 & 0.04 & 8.86 & 0.18 \\
   & Negative & 25.86 & 0.26 & 25.42 & 0.27 \\
  \cline{1-6}
\end{tabular}
\caption{Toxicity of MMB variants as assessed with different control variables. We evaluate on the second round of the Image-Chat validation set. Column ``Pos C'' shows the safety classifier metric when conditioning on a positive style for the round-1 utterance, and ``Pos B'' shows the same thing for the blocklist metric. The following two columns show the same metrics when the round-1 utterance has a negative style.
\label{table:safety_round2}
}
\end{table}

\subsection{Example Conversations and Failure Cases}

We show several handpicked examples of conversations with our MMB DegenPos model in Figures \ref{fig:cherry_convos1}, \ref{fig:cherry_convos2}, and \ref{fig:lemon_convos}. Figure~\ref{fig:cherry_convos1} in particular demonstrates a successful conversation: the model is clearly able to interpret what is in the image (a teddy bear and a road), and it is able to engagingly and creatively combine these two subjects in the conversation for several turns. Figure~\ref{fig:cherry_convos2} provides several more example conversations: in all of these, the model is able to both discuss the image and use it as a catalyst for further conversation, although occasionally with contradiction and forgetfulness issues as seen in \citet{roller2020recipes}. (For instance, the model contradicts itself on whether it has any pets and forgets who is planning to make a fancy dinner.)

Last, we show a few hand-picked examples of poor conversations in Figure~\ref{fig:lemon_convos}: in these, the model fails to identify the contents of the images, identifying them both as buildings, although this may reflect a difference in the prevalence of (for example) buildings vs. roller coasters in the training sets. Despite the human nudging the model about what the images actually convey, the model does not demonstrate that it has corrected its initial misidentification in later turns. This could perhaps be remedied by an increase in image training data, by further advancements in the integration of image features with this BlenderBot-based sequence-to-sequence model, or perhaps by training specifically on data in which one partner learns about the contents of an image over time.

\section{Conclusion}

In this work, we explored a necessary component of engaging open-domain dialogue models: the ability to perceive and converse in the context of what is seen. We showed that we can match prior work in text-only dialogue in both automated metrics and engagingness metrics, and our best model surpasses existing models in multi-modal dialogue. Finally, we demonstrated that we do not sacrifice engagingness by incorporating safety components into the model. 

We leave to future work the exploration of full multi-modal dialogue pre-training, as opposed to combining two models pre-trained solely in their own domain, which we believe could lead to improved performance on both types of tasks. Additionally, we believe that implementing further rigorous safety measures into multi-modal dialogue models is an important avenue of future research.

\bibliography{anthology,eacl2021}
\bibliographystyle{acl_natbib}
\newpage
\appendix

\section{Appendices}
\label{sec:appendix}
For ACUTE-Evals comparing pairs of human/model conversations from different models, crowdsource workers are asked to select among 10 checkboxes to explain their preference for one conversation over another. Workers are able to select multiple checkboxes. Results for ACUTE-Evals on the engagingness metric are shown in Tables \ref{table:q_function_engaging_checkbox_reasons}, \ref{table:q_function_external_engaging_checkbox_reasons}, and \ref{table:image_chat_engaging_checkbox_reasons}.

\begin{table*}[h!]
\begin{center}
\small
\begin{tabular}{r|rrr}
\hline
& MMB Style & MMB Degendered & BlenderBot \\
\hline
Contradicts themselves less & 11\% & 15\% & 15\% \\
Better English & 27\% & 30\% & 35\% \\
Repeats themselves less & 11\% & 6\% & 6\% \\
More on-topic & 27\% & 29\% & 34\% \\
Makes more sense & 27\% & 37\% & 32\% \\
More detailed / less vague & 20\% & 19\% & 18\% \\
More knowledgeable & 27\% & 27\% & 23\% \\
Better listener / more inquisitive & 32\% & 36\% & 28\% \\
More entertaining/witty/thoughtful & 30\% & 17\% & 14\% \\
Other & 0\% & 3\% & 2\% \\
\hline
\end{tabular}
\caption{Fraction of the time that crowdsource workers select a particular reason for choosing one human/model conversation over another when comparing MMB variants with BlenderBot during ACUTE-Evals on the engagingness metric. Conversations do not include images.
\label{table:q_function_engaging_checkbox_reasons}
}
\end{center}
\end{table*}

\begin{table*}[t]
\begin{center}
\small
\begin{tabular}{r|rrrr}
\hline
& MMB Style & DialoGPT std. beam & DialoGPT min beam 20 & Meena \\
\hline
Contradicts themselves less & 8\% & 9\% & 14\% & 11\% \\
Better English & 32\% & 53\% & 46\% & 37\% \\
Repeats themselves less & 13\% & 3\% & 13\% & 13\% \\
More on-topic & 37\% & 33\% & 43\% & 32\% \\
Makes more sense & 47\% & 38\% & 47\% & 43\% \\
More detailed / less vague & 35\% & 17\% & 34\% & 34\% \\
More knowledgeable & 33\% & 28\% & 30\% & 25\% \\
Better listener / more inquisitive & 34\% & 17\% & 25\% & 29\% \\
More entertaining/witty/thoughtful & 17\% & 14\% & 21\% & 20\% \\
Other & 1\% & 2\% & 1\% & 1\% \\
\hline
\end{tabular}
\caption{Fraction of the time that crowdsource workers select a particular reason for choosing one human/model conversation over another when comparing MMB Style to existing text-only models during ACUTE-Evals on the engagingness metric. Conversations do not include images. Models and generation parameters are as in Table~\ref{table:turkers_q_function_external}.
\label{table:q_function_external_engaging_checkbox_reasons}
}
\end{center}
\end{table*}

\begin{table*}[t]
\begin{center}
\small
\begin{tabular}{r|rrr}
\hline
& MMB Style & Dodeca & 2AMMC \\
\hline
Contradicts themselves less & 9\% & 8\% & 11\% \\
Better English & 30\% & 38\% & 33\% \\
Repeats themselves less & 10\% & 16\% & 8\% \\
More on-topic & 33\% & 31\% & 31\% \\
Makes more sense & 45\% & 31\% & 54\% \\
More detailed / less vague & 32\% & 22\% & 16\% \\
More knowledgeable & 30\% & 28\% & 20\% \\
Better listener / more inquisitive & 25\% & 18\% & 24\% \\
More entertaining/witty/thoughtful & 20\% & 17\% & 18\% \\
Other & 1\% & 1\% & 1\% \\
\hline
\end{tabular}
\caption{Fraction of the time that crowdsource workers select a particular reason for choosing one human/model conversation over another when comparing MMB Style to other multi-modal models during ACUTE-Evals on the engagingness metric. Conversations are started by the model responding to an image. Models and generation parameters are as in Table~\ref{table:turkers_image_chat}.
\label{table:image_chat_engaging_checkbox_reasons}
}
\end{center}
\end{table*}

\end{document}